%% file: acl_latex.tex
\documentclass[11pt]{article}

\usepackage[]{acl}

\usepackage{times}
\usepackage{latexsym}

\usepackage[T1]{fontenc}

\usepackage[utf8]{inputenc}

\usepackage{microtype}

\usepackage{inconsolata}

\usepackage{graphicx}

\usepackage{fontawesome5}
\newcommand{\ours}{\textit{\textbf{TIDE}}}
\usepackage{amsmath}
\usepackage{amsthm}
\usepackage{amsfonts}
\usepackage{amssymb}
\usepackage{multirow}
\usepackage{multicol}
\usepackage{booktabs}
\usepackage{xcolor}
\usepackage{colortbl}
\usepackage{tcolorbox}
\tcbuselibrary{skins, breakable}
\usepackage{algorithm}
\usepackage{algorithmic}
\usepackage{framed}
\usepackage{listings}
\usepackage{xcolor}
\definecolor{iceblue}{HTML}{E6F2FF}  
\definecolor{deepblue}{HTML}{0073E6} 
\definecolor{LakeBlue}{RGB}{0,61,153}

\tcbuselibrary{breakable}

\definecolor{mycellcolor}{HTML}{FFFFFF}
\definecolor{mycellcolor2}{HTML}{e6f4f1}
\definecolor{veronica-red}{RGB}{196,30,58}

\author{
Hang Yan\textsuperscript{$\diamondsuit$*} \quad
Xinyu Che\textsuperscript{$\diamondsuit$*} \quad
Fangzhi Xu\textsuperscript{$\diamondsuit$*$\dagger$} \quad
Qiushi Sun\textsuperscript{$\heartsuit$} \quad \\
\bf{
Zichen Ding\textsuperscript{$\spadesuit$} \quad
Kanzhi Cheng\textsuperscript{$\triangle$} \quad
Jian Zhang\textsuperscript{$\diamondsuit$} \quad
Tao Qin\textsuperscript{$\diamondsuit$} \quad
Jun Liu\textsuperscript{$\diamondsuit$} \quad
Qika Lin\textsuperscript{$\clubsuit$$\dagger$} \quad
}\\
\textsuperscript{$\diamondsuit$}Xi'an Jiaotong University 
\textsuperscript{$\heartsuit$}The University of Hong Kong \\
\textsuperscript{$\spadesuit$}Shanghai AI Laboratory
\textsuperscript{$\triangle$}Nanjing University
\textsuperscript{$\clubsuit$}National University of Singapore\\
\texttt{hyan@stu.xjtu.edu.cn} \:
\texttt{fangzhixu98@gmail.com} \:
\texttt{qikalin@foxmail.com}
\\
* means equal contribution \quad
$\dagger$ denotes corresponding authors \\
\faGithub\ \url{https://github.com/yayayacc/TIDE}
}

\title{%
\hspace*{0.7em}%
\begin{minipage}{0.12\textwidth}
  \centering
  \includegraphics[height=2\baselineskip]{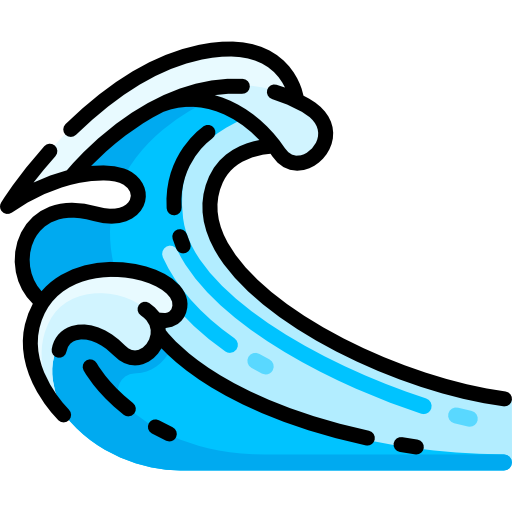}
\end{minipage}
\hspace{-1em}
\begin{minipage}{0.8\textwidth}
  \raggedright
  \textbf{TIDE: Trajectory-based Diagnostic Evaluation of\\
  \qquad Test-Time Improvement in LLM Agents}
\end{minipage}
}

\begin{document}
\maketitle

\begin{abstract}
Recent advances in autonomous LLM agents demonstrate their ability to improve performance through iterative interaction with the environment. 
We define this paradigm as \textbf{Test-Time Improvement (TTI)}.
However, the mechanisms under how and why TTI succeed or fail remain poorly understood, and existing evaluation metrics fail to capture their task optimization efficiency, behavior adaptation after erroneous actions, and the specific utility of working memory for task completion.
To address these gaps, we propose \textbf{T}est-time \textbf{I}mprovement \textbf{D}iagnostic \textbf{E}valuation (\ours), an agent-agnostic and environment-agnostic framework that decomposes TTI into three comprehensive and interconnected dimensions. The framework measures (1) the overall temporal dynamics of task completion and (2) identifies whether performance is primarily constrained by recursive looping behaviors or (3) by burdensome accumulated memory.
Through extensive experiments across diverse agents and environments, \ours~highlights that improving agent performance requires more than scaling internal reasoning, calling for explicitly optimizing the interaction dynamics between the agent and the environment.
\end{abstract}

\input{Sections/1Intro}

\input{Sections/2Preliminaries}

\input{Sections/3Methodology}

\input{Sections/4Framework}
\input{Sections/5Relatedwork}

\input{Sections/6Conclustion}

\bibliography{custom}

\clearpage
\newpage
\appendix

\input{Sections/AppendixA}

\input{Sections/AppendixB}

\input{Sections/AppendixC}

\end{document}

%% file: Sections/1Intro.tex
\section{Introduction}

Through active interaction with the environment, LLM agents~\cite{yao2023reactsynergizingreasoningacting, xi2025rise} demonstrate substantial potential for handling complex real-world tasks. 
This capability has enabled a wide range of practical applications, including coding agents~\citep{wang2024executable, chen2025locagentgraphguidedllmagents} and GUI agents~\citep{qin2025uitarspioneeringautomatedgui, wu2024osatlasfoundationactionmodel}.

\begin{figure*}[t]
  \centering
  \includegraphics[width=2\columnwidth]{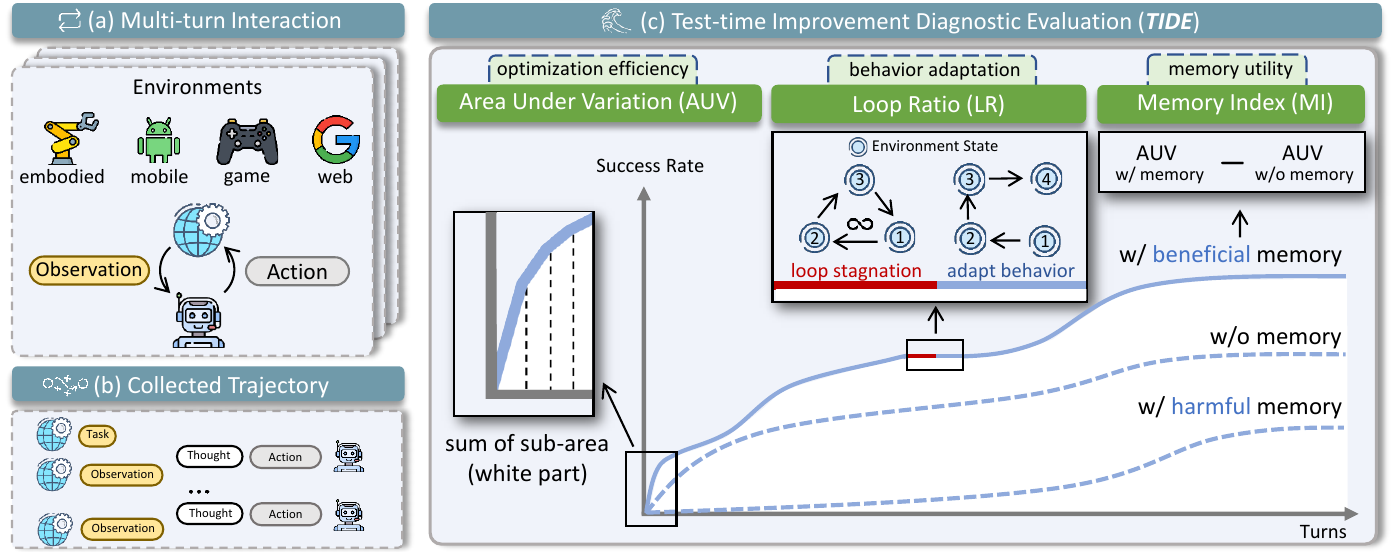}
  \caption{Overview of our trajectory-based diagnostic evaluation framework. (a) An agent completes tasks through multi-turn interaction with the environment. (b) Interaction trajectories are collected for diagnostic analysis. (c) \ours~provides a unified and interconnected diagnosis of TTI trajectories via three complementary metrics. AUV quantifies optimization efficiency by aggregating trapezoidal sub-areas along the trajectory; LR distinguishes loop-induced stagnation from behavioral adaptation; MI isolates and analyzes the contribution of working memory.}
  \label{fig:intro}
\end{figure*}

Due to the complexity and unpredictability of real-world environments~\citep{jimenez2024swebenchlanguagemodelsresolve, zhou2024webarenarealisticwebenvironment, wang2025mmbench}, relying solely on internal reasoning is often insufficient for optimal task completion~\citep{parisi2024beyond}. 
This challenge necessitates a critical adaptive capacity, in which the agent accumulates experience through continuous interaction to iteratively rectify its actions.
We define this dynamic improvement process as \textbf{Test-Time Improvement (TTI)}.
Despite the critical role of TTI for agent autonomy, a rigorous understanding of how such improvement unfolds, stagnates, or deteriorates remains a missing piece.
We argue that TTI reflects the interplay of three fundamental aspects of interactive optimization: \textbf{how efficiently} an agent converts interaction budget into progress, \textbf{how adaptively} it responds to errors and feedback, and \textbf{how effectively} it leverages the accumulated interaction history. 
These three aspects, as shown in Figure~\ref{fig:intro}(c), are inherently interconnected and together determine whether test-time interaction leads to genuine improvement.
Guided by this decomposition, we structure our study around the following three core research questions.

Firstly, beyond eventual success, a competent agent is expected to improve efficiently and progressively through interaction with the environment.
However, widely adopted static metrics, such as Success Rate (SR)~\citep{ wang2024executable,xu2025theagentcompanybenchmarkingllmagents}, collapse the informative trajectory into a single binary outcome, treating an efficient one-step success as equivalent to a delayed success after extensive exploration.
Based on this observation, 
we propose 
RQ I: \textit{How to quantify the optimization efficiency of an agent's performance evolution?}

Secondly, errors are inevitable in complex environments, and effective agents must demonstrate behavior adaptation that corrects their actions by learning from failures. However, existing metrics, such as the number of interaction turns~\cite{zhang2025efficientllmgroundingembodied, pitre-etal-2025-consensagent}, are largely content-agnostic, conflating genuine corrective behavior adaptation with repetitive failure actions.
As a result, an agent may appear active while repeatedly executing ineffective strategies.
This prompts RQ II: \textit{How can we formalize the boundary between behavior adaptation and recursive failure?}

Thirdly, test-time interaction inevitably accumulates working memory that may contain both useful experiences and misleading noise. 
While longer contexts are often assumed to be beneficial, their actual impact on decision quality remains unclear, as existing analyses~\citep{qiu2025locobench,chhikara2025mem0} typically conflate memory effects with other confounding factors, such as model scale or interaction length.
Accordingly, we propose RQ~III: \textit{How to quantify the utility of accumulated interaction memories to agent performance?}

To systematically address the research questions, we introduce Test-time Improvement Diagnostic Evaluation (\ours), a lightweight, agent-agnostic and environment-agnostic framework designed to diagnose how LLM agents evolve during interaction.
Specifically, \ours~formalizes TTI via three metrics corresponding to the three above-mentioned RQs.
AUV (Area Under Variation) quantifies optimization efficiency by capturing how quickly and steadily an agent succeeds over interaction.
LR (Loop Ratio) measures behavior stagnation by identifying loop patterns, enabling the distinction between effective behavior adaptation and recursive failure.
MI (Memory Index) isolates the utility of accumulated interaction history, quantifying how working memory contributes to performance.
Together, these metrics provide a unified diagnostic view of TTI: AUV quantifies overall temporal efficiency, while LR and MI pinpoint its bottlenecks as recursive loops and memory burdens. 
Through comprehensive experiments, we highlight the need to move beyond scaling internal reasoning alone, toward explicitly optimizing the dynamics of agent–environment interaction. 

We highlight our main contributions as follows:

\noindent (1) \textbf{Conceptualization of Test-Time Improvement for LLM agents.}
We first formalize Test-time Improvement as a multi-dimensional, interaction-driven process beyond accuracy.

\noindent (2) \textbf{Diagnostic Evaluation Framework.}
We introduce \ours, a trajectory-based diagnostic evaluation framework decomposing TTI into optimization efficiency, behavior adaptation, and memory utility.

\noindent (3) \textbf{Empirical Diagnosis Across Environments.}
Extensive experiments reveal failure modes and bottlenecks that are invisible to existing metrics, providing actionable insights for optimizing interactive LLM agents.

%% file: Sections/2Preliminaries.tex
\section{Preliminaries}
\label{sec:pre}
\subsection{Multi-turn Interaction Agent}
We formalize the multi-turn interaction between the LLM agent and the environment as a Partially Observable Markov Decision Process (POMDP), defined by the tuple $\mathcal{M} = \langle \mathcal{S}, \mathcal{A}, \mathcal{O}, \mathcal{F}, \mathcal{R}, g \rangle$.  $\mathcal{S}$ represents the latent state space. $\mathcal{A}$ denotes the action space. $\mathcal{O}$ is the observation space. The transition function $\mathcal{F}: \mathcal{S} \times \mathcal{A} \rightarrow \mathcal{S}$. The interaction process yields a rollout $\tau = [o_0, a_0, o_1, a_1, \dots, o_T]$, where $a \in \mathcal{A}$ and $o \in \mathcal{O}$. $\mathcal{R}(\tau)\rightarrow \{0, 1\}$ indicates whether the trajectory $l$ reaches the goal $g$. Additionally, MDP serves as a special case of POMDP, where the unobservable state space is empty.
\input{Tables/auv}
\subsection{Evaluation Protocols}
Our experimental setup is categorized into \textbf{reasoning-bound} (MDP) and \textbf{information-bound} (POMDP) according to their dependency on external feedback. For reasoning-bound tasks, the solution trajectory can be deduced independently of immediate environmental responses. Conversely, for information-bound tasks, the agent must actively interact with the surroundings to acquire information.
Detailed setup is in Appendix~\ref{appendix:environments}.


%% file: Tables/auv.tex
\begin{table*}[t]
\centering
\footnotesize
\resizebox{\linewidth}{!}{
\begin{tabular}{l|cc|cc|cc|cc|cc}
\toprule
\multirow{2}{*}{\textbf{Model}}  &
\multicolumn{2}{c|}{\textbf{BlocksWorld}} &
\multicolumn{2}{c|}{\textbf{FrozenLake}} &
\multicolumn{2}{c|}{\textbf{Sudoku}} &
\multicolumn{2}{c|}{\textbf{AlfWorld}} &
\multicolumn{2}{c}{\textbf{WebShop}} \\
& \textbf{SR$\uparrow$} & \textbf{AUV$\uparrow$} 
& \textbf{SR$\uparrow$} & \textbf{AUV$\uparrow$}
& \textbf{SR$\uparrow$} & \textbf{AUV$\uparrow$}
& \textbf{SR$\uparrow$} & \textbf{AUV$\uparrow$}
& \textbf{SR$\uparrow$} & \textbf{AUV$\uparrow$} \\
\midrule
\multicolumn{11}{c}{\cellcolor{gray!15} Non-thinking Model}  \\
\midrule
\cellcolor{mycellcolor}Qwen3-4B-Instruct & 40.0 & 30.8 & 59.0 & 46.1 & 29.0 & 19.3 & 52.1 & 38.3 & 15.0 & 9.6 \\
\cellcolor{mycellcolor}Qwen3-30B-A3B-Instruct & 64.0 & 45.4 & 61.0 & 49.0 & 77.0 & 48.8 & 62.1 & 48.1 & 20.8 & 13.7 \\
\cellcolor{mycellcolor}Llama-3.1-8B-Instruct & 32.0 & 21.9 & 12.0 & 8.4 & 4.0 & 2.1 & 16.4 & 12.8 & 20.8 & 13.1  \\
\cellcolor{mycellcolor}Llama-3.3-70B-Instruct & 96.0 & 68.2 & 44.0 & 34.7 & 40.0 & 24.9 & 59.3 & 46.7 & 26.8 & 17.6 \\
\cellcolor{mycellcolor}GLM-4-9B-Chat & 23.0 & 18.2 & 10.0 & 7.2 & 4.0 & 2.0 & 25.7 & 21.8 & 16.4 & 11.5 \\
\cellcolor{mycellcolor}GLM-4-32B-0414 & 83.0 & 57.5 & 67.0 & 49.9 & 55.0 & 34.4 & 37.9 & 29.0 & 25.4 & 17.2 \\
\cellcolor{mycellcolor}Mistral-7B-Instruct & 7.0 & 5.0 & 6.0 & 4.6 & 1.0 & 0.4 & 8.6 & 7.6 & 6.6 & 4.4 \\
\cellcolor{mycellcolor}Ministral-3-14B-Instruct & 72.0 & 39.0 & 56.0 & 35.8 & 18.0 & 13.3 & 22.1 & 15.8 & 15.2 & 5.4 \\
\cellcolor{mycellcolor}Phi-4 & 56.0 & 41.2 & 66.0 & 50.8 & 33.0 & 21.4 & 25.0 & 18.6 & 18.8 & 12.2 \\
\cellcolor{mycellcolor}DeepSeek-V3.2 & 98.0 & 71.1 & 97.0 & 73.1 & 93.0 & 56.8 & 80.7 & 59.0 & 41.0 & 24.8 \\
\rowcolor{mycellcolor2}
Gemini 2.5 Flash & 99.0 & 72.0 & 97.0 & 70.7 & 97.0 & 60.4 & 78.6 & 58.1 & 35.0 & 21.6 \\
\midrule
\multicolumn{11}{c}{\cellcolor{gray!15} Thinking Model}  \\
\midrule
\cellcolor{mycellcolor}Qwen3-4B-Thinking & 77.0 & 47.9 & 91.0 & 66.7 & 55.0 & 30.7 & 57.1 & 40.9 & 12.2 & 8.2 \\
\cellcolor{mycellcolor}Qwen3-30B-A3B-Thinking & 98.0 & 69.8 & 93.0 & 68.0 & 95.0 & 55.6 & 66.4 & 50.7 & 21.0 & 14.6 \\
\cellcolor{mycellcolor}Phi-4-reasoning & 68.0 & 51.3 & 62.0 & 48.8 & 51.0 & 37.4 & 15.7 & 9.8 & 22.6 & 14.3 \\
\cellcolor{mycellcolor}gpt-oss-120b & \textbf{100.0} & 73.7 & 99.0 & 73.8 & 97.0 & \textbf{60.2} & 50.7 & 29.0 & 26.2 & 11.1 \\
\cellcolor{mycellcolor}DeepSeek-R1 & \textbf{100.0} & 74.9 & 96.0 & 72.6 & 95.0 & 59.4 & \textbf{83.6} & \textbf{64.1} & 38.4 & 24.1 \\
\rowcolor{mycellcolor2}
Gemini 2.5 Pro & \textbf{100.0} & \textbf{75.2} & \textbf{100.0} & \textbf{75.0} & \textbf{99.0} & \textbf{60.2} & 80.7 & 62.9 & \textbf{43.2} & \textbf{27.1} \\
\bottomrule
\end{tabular}
}
\caption{We report overall success rate (SR) and area under variation (AUV) on 5 widely adopted benchmarks. $\uparrow$ means a higher value is better. \textcolor[HTML]{289BA2}{Colored model name} represents proprietary models. Best values are in bold.}
\label{tab:auv}
\end{table*}

%% file: Sections/3Methodology.tex
\section{Breakdown Test-Time Improvement}
\label{sec:method}
In this section, we deconstruct TTI through three agent-agnostic and environment-agnostic indicators, additionally providing insights through comprehensive experiments.



\subsection{Optimization Efficiency Diagnostic: Modeling Temporal Dynamics }
\label{sec:wcr}

To quantify the optimization efficiency of an agent's performance evolution, we model agent performance as a discrete-time evolution process over interaction steps. Instead of assessing an interaction trajectory through a terminal outcome, we characterize TTI progress with a variation curve $P_t$, which denotes the cumulative proportion of tasks successfully solved within the first $t$ interaction turns. 
Tracking $P_t$ over $t$ explicitly captures how quickly and steadily an agent converts additional interaction budget into task success, enabling a principled temporal analysis of the task completion dynamics.

Building upon $P_t$, we introduce a holistic evaluation metric, denoted as Area Under Variation (AUV), which is defined as the integral area under the performance variation curve.
Formally, let $[0, t_{max}]$ denote the domain-specific evaluation window, where $t_{max}$ is the experimental evaluation horizon in each environment. Detailed configuration for $t_{max}$ is  in Appendix~\ref{appendix:t_min_max}. AUV is calculated within this evaluation window as follows:



\begin{equation}
    \text{AUV} = \frac{1}{t_{max}}\sum_{t=0}^{t_{max}-1}  \frac{P_t + P_{t+1}}{2}, 
    \label{eq:auv}
\end{equation}
where the trapezoidal sum $\frac{P_t + P_{t+1}}{2}$ accounts for the incremental performance gain between successive steps.
The denominator $t_{max}$ serves as a normalization constant, ensuring that the metric is bounded between 0 (no improvement) and 1 (instantaneous task completion). 

More theoretical proofs provided in Appendix~\ref{appendix:auv_theory} indicate that AUV captures more details than SR.
We report AUV scores along with the corresponding SR in Table~\ref{tab:auv} and reveal the following findings:

\begin{figure}[t]
  \includegraphics[width=\columnwidth]{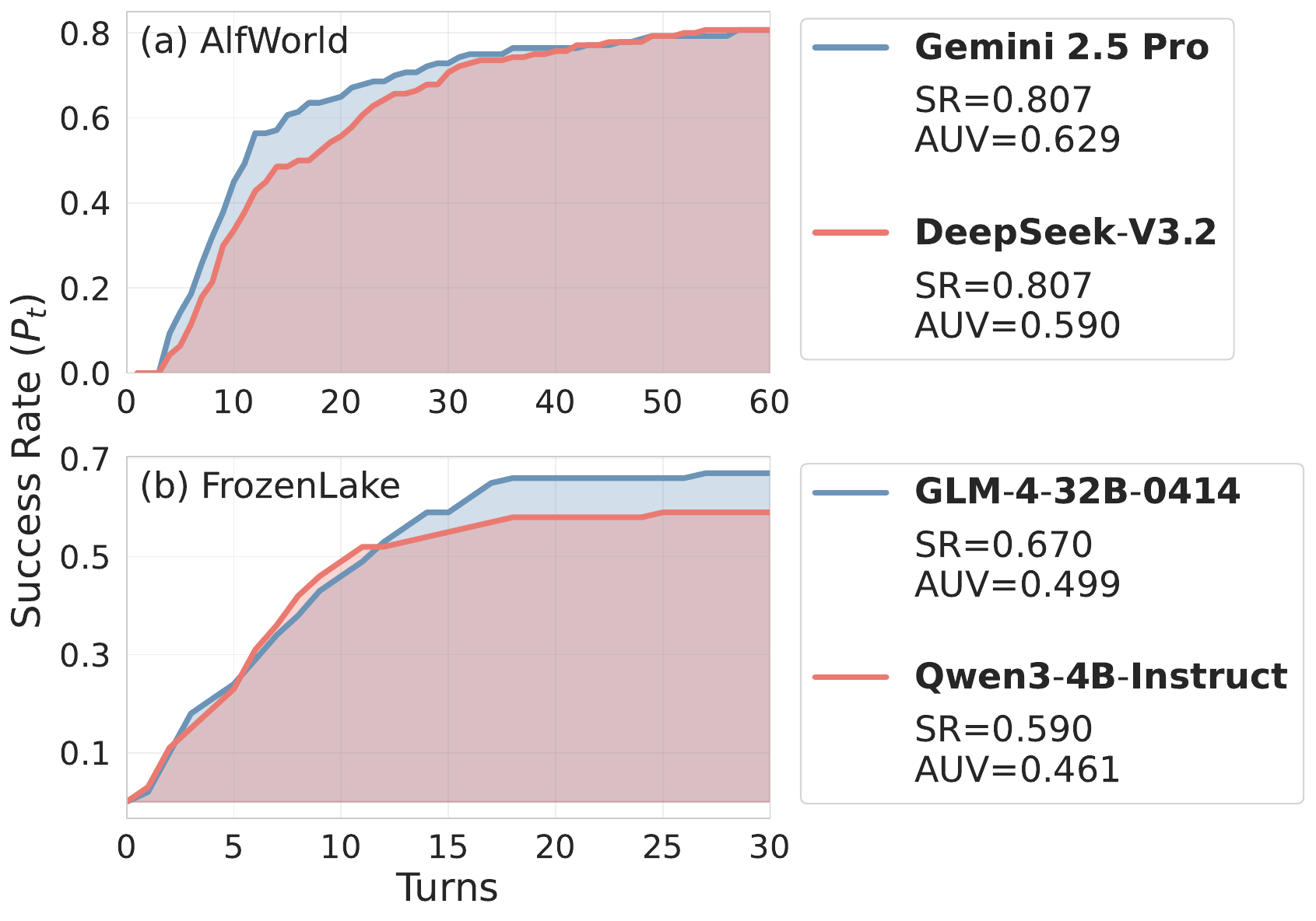}
  \caption{SR curves on three environments. For comparison, we report both AUV and SR. We reveal that SR obscures underlying efficiency differences, and few interaction turns fail to exhibit the agent's TTI capability.}
  \label{fig:passT}
\end{figure}

\paragraph{AUV Quantifies Temporal Efficiency.}
As illustrated in Figure~\ref{fig:passT}(a), while \textit{DeepSeek-V3.2} and \textit{Gemini 2.5 Pro} converge to nearly identical final success rates 0.807 in AlfWorld, they exhibit a notable divergence in AUV. Specifically, \textit{Gemini 2.5 Pro} achieves a higher AUV 0.629, indicating early-stage TTI efficiency. This indicates that AUV successfully captures the distinguishing temporal efficiency of different agents. Notably, in simple environments where SR saturates at 1 for different models, such as BlocksWorld, AUV continuously demonstrates unique value. AUV captures a fine-grained distinction of optimization efficiency, which cannot be discovered by SR.

\paragraph{AUV Rewards Sustained Convergence Dynamics.}
As illustrated in Figure~\ref{fig:passT}(b), although \textit{GLM-4-32B-0414} exhibits a comparable or even lower SR in the early stages compared to \textit{Qwen3-4B-Instruct}, it achieves significantly higher performance in later turns, demonstrating a superior TTI capability. 
Crucially, the substantial marginal gains realized in the later stages result in \textit{GLM-4-32B-0414} achieving a superior overall AUV 0.499. 
This confirms that AUV effectively rewards agents that possess the continuous improvement capability to solve complex problems.

In particular, although SR and AUV reveal similar evaluation results in this case, we argue that AUV is not equivalent to SR. 
More experimental details can be found in Appendix~\ref{appendix:auv_validation}.

\paragraph{TTI Success Depends on Agent-Environment Match.}
TTI efficacy is not a universal capability but is subject to agent-environment constraints. For instance, \textit{Llama3.3-70B-Instruct} surpasses \textit{Qwen3-30B-A3B-Instruct} in BlocksWorld, a trend that sharply reverses in FrozenLake and Sudoku.
This highlights the imperative for holistic evaluation to accurately verify a model's generalized TTI capability, rather than its specialized performance.



\begin{tcolorbox}[
    colback=iceblue,
    colframe=deepblue,
    arc=3mm,
    boxrule=0.8pt,
    left=6pt, right=6pt, top=6pt, bottom=6pt
]
\noindent \textbf{Takeaway 1:} Current metrics overlook the temporal dynamics of TTI. Moreover, TTI efficacy depends on agent-environment match rather than purely intrinsic.
\end{tcolorbox}

\subsection{Behavior Adaptation Diagnostic: Characterizing Recursive Loops}
\label{sec:lr}

To formalize the boundary between behavior adaptation and recursive failure, we introduce an auto-detection algorithm through interpreting an agent’s interaction trajectory as a path over the latent environment state space (described in Sec~\ref{sec:pre}). 
From this perspective, actions correspond to directed transitions between states, allowing the trajectory to be analyzed as a graph-structured process. Then we rigorously distinguish adaptive exploration from degenerate repetition by identifying recursive cycle structures in the trajectory-based graph, which serve as a signal of adaptation failure.

We define the cycle as $l_{ij} = [s_i, a_i,..., a_{j-1}, s_j]$, where $s_i = s_j$ and $i \neq j$. The environment state departs from a node $s_i$ and eventually returns to the exact same node $s_j$, yielding no goal progress. The set of these cycle units is denoted as $\mathcal{L}_{cycle}$. Notably, we impose a non-recursive constraint that each cycle does not contain any nested sub-cycles to avoid ambiguous decomposition. Moreover, a single action resulting in no change of the environment state is also denoted as a cycle:
    
\begin{equation}
\begin{aligned}
\mathcal{L}_{cycle} = \{\, &l_{ij} \mid\; 
     s_j = s_i,\; i < j,\\
    &  \text{and } \forall i \le p <q<j,s_p \neq s_q  \,\}.
\end{aligned}
\end{equation}

Once encountering a cycle, the agent can utilize it to traverse a new path or repeat the previous cycle without task progress. The latter loop behavior significantly diminishes exploration without task progress.
We define a subset of $\mathcal{L}_{cycle}$ as $\mathcal{L}_{\text{loop}}$, where the agent repeats the previous cycle and the two cycles should occur consecutively:
\begin{equation}
\begin{aligned}
\mathcal{L}_{\text{loop}}
    = \{\, l_{jk} \in \mathcal{L}_{\text{cycle}} 
        \mid l_{ij} = &l_{jk},\; i < j < k,\; \\
        &l_{ij} \in \mathcal{L}_{\text{cycle}} \,\}.
\end{aligned}
\end{equation}

We then quantify this loop behavior by Loop Ratio (LR)  as the proportion of these redundant loop actions relative to the total actions:
\begin{equation}
\label{eq:lr}
\text{LR} = \frac{\sum_{l_{ij} \in \mathcal{L}_{\text{loop}}} j-i}{\text{Total Actions}},
\end{equation}
where a lower LR suggests less stagnation, signifying that the agent actively attempts to alter its strategy upon failure rather than succumbing to degenerate repetition.
\input{Tables/lr}

More implementation details and the auto-detection algorithm can be found in Appendix~\ref{appendix:lr}.
We demonstrate the LR results in Table~\ref{tab:lr} and reveal the following findings:

\begin{figure}[t]
  \includegraphics[width=\columnwidth]{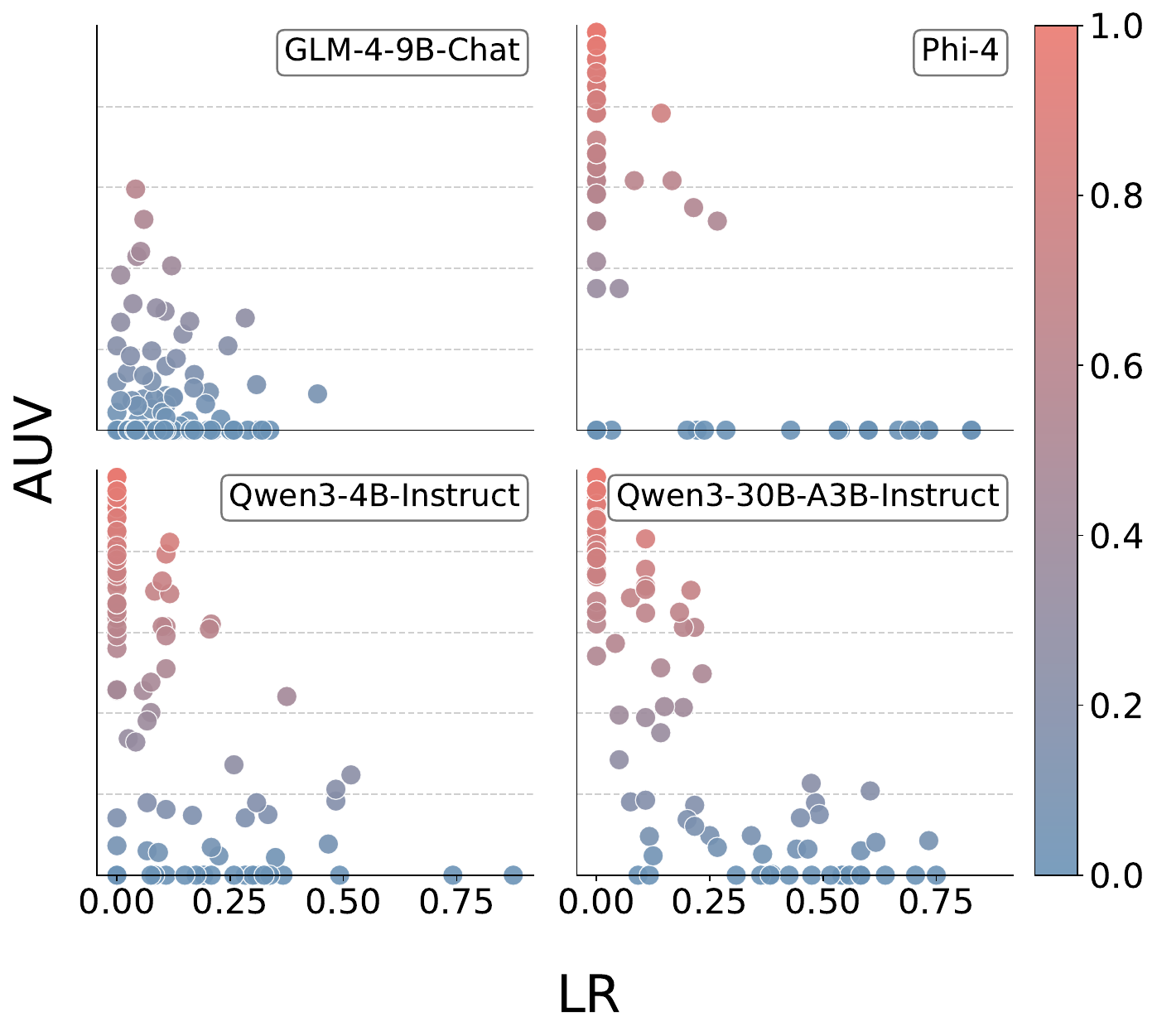}
  \caption{The relationship between Loop Ratio and corresponding AUV for each task in FrozenLake.}
  \label{fig:lr_auv}
\end{figure}

\paragraph{Loop is Widely Observed in Current Models.}
A majority of the evaluated LLM agents exhibit remarkably high LR. This indicates that instead of adapting their behaviors based on negative environment feedback, the models tend to persist in a state of recursive failure, repeatedly attempting the same erroneous actions. 
For instance, \textit{Qwen3-4B-Instruct} records a LR of 32.0\% in the FrozenLake environment. 
Such rigidity suggests that these agents suffer from a deficiency in valid behavior adaptation.

More details demonstrating that the loop phenomenon is significantly associated with over-confidence can be found in Appendix~\ref{appendix:loop_analysis}.


\paragraph{High LR Indicates Suboptimal TTI.} To investigate the extent to which the loop phenomenon constrains performance, we visualize the relationship between LR and AUV in Figure~\ref{fig:lr_auv}. Every data point represents a task in each environment. We observe a statistically significant inverse relationship across four models, where a high LR is associated with a low AUV. This suggests that loops are strongly associated with suboptimal TTI. Notably, as indicated by the data points in the lower-left corner of each sub-figure, a low Loop Ratio does not strictly guaranty a high AUV. Thus, we conclude that minimizing loops is a necessary but not sufficient condition for high TTI capacity.

\paragraph{Scaling Mitigates Stagnation.}
As shown in Figure~\ref{tab:lr}, within the same model family, larger models consistently demonstrate significantly lower LR. For instance, scaling from \textit{Qwen3-4B-Instruct} to \textit{Qwen3-30B-A3B-Instruct} mitigates the average LR from 15.8\% to 1.0\% in BlocksWorld, with an even more significant reduction in WebShop, from 36.7\% to 5.7\%. This inverse correlation suggests that an increase in the parameter endows agents with greater strategic diversity and behavior adaptation capabilities. Moreover, extremely large models, such as the \textit{DeepSeek} and \textit{Gemini} series, consistently demonstrate negligible LR across the five environments.


\begin{tcolorbox}[
    colback=iceblue,
    colframe=deepblue,
    arc=3mm,
    boxrule=0.8pt,
    left=6pt, right=6pt, top=6pt, bottom=6pt
]
\noindent \textbf{Takeaway 2:} Many advanced models frequently succumb to stubborn loops associated with overconfidence. Moreover, low LR is a necessary condition for optimal TTI.
\end{tcolorbox}

\subsection{Memory Utility Diagnostic: Assessing Working Memory Contribution}
\label{sec:mri}

To quantify and analyze the utility of interaction memories to agent performance, we adopt an ablation evaluation protocol that compares agent performance with and without access to the accumulated interaction trajectory. By keeping the agent and environment fixed and varying only the availability of past trajectories, this setup enables a direct quantification of the specific contribution of working memory to optimization dynamics.


Building on this setup, we define the Memory Index (MI) as a quantitative measure of this effect, computed as the performance gap between the w/ memory and w/o memory configurations, thereby capturing the specific contribution of the expanding interaction memory:
\begin{equation}
\label{eq:mri}
\text{MI} = \text{AUV}_{\text{w/ memory}} - \text{AUV}_{\text{w/o memory}},
\end{equation}
where $\text{AUV}_{\text{w/ memory}}$ denotes the performance using complete working memory, and $\text{AUV}_{\text{w/o memory}}$ represents the performance when the agent is restricted to only the task description and immediate observation, without access to earlier trajectory.

A positive MI reflects a scenario in which working memory relates positively to TTI, while a negative MI indicates a harmful influence.

We emphasize that MI is designed to measure the efficacy of working memory rather than to depict the agent's competence in performing memory management. For instance, an agent with both $\text{AUV}_{\text{w/ memory}}=1$ and $ \text{AUV}_{\text{w/o memory}}=1$ produces a zero MI, but this does not imply a failure in memory management. It suggests that the memory information plays a negligible role between these settings. We demonstrate the MI results in Figure~\ref{fig:mri} and reveal the following findings:
\begin{figure}[t]
  \includegraphics[width=\columnwidth]{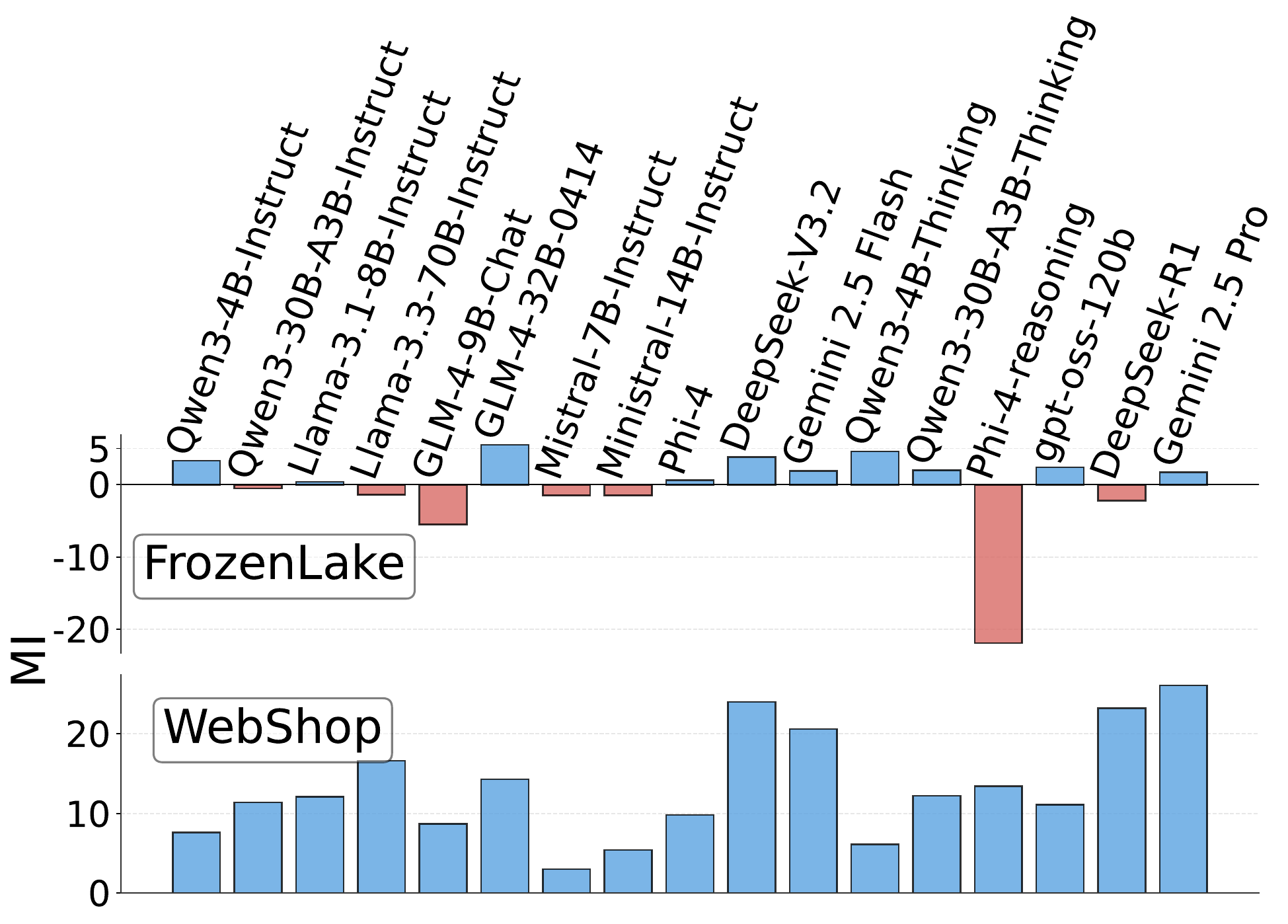}
  \caption{We report MI in FrozenLake and WebShop. More MI results can be found in Appendix~\ref{appendix:mri}.}
  \label{fig:mri}
\end{figure}

\paragraph{Negative Memory Influence is Widely Observed.}
As illustrated in Figure~\ref{fig:mri}, 
contrary to the prevailing assumption that expanding context promotes agent performance, our MI data reveal a counter-intuitive phenomenon: as evidenced by negative memory influence of multiple models in FrozenLake, simply leveraging working memory without further management often acts as a cognitive burden in reasoning-bound environments, especially for open-source models. This may  because excessive irrelevant details in memory hinder reasoning, suggesting that a simple scaling context length is not a universally valid strategy for performance enhancement, highlighting the need for active memory management.
\begin{figure}[t]
  \includegraphics[width=\columnwidth]{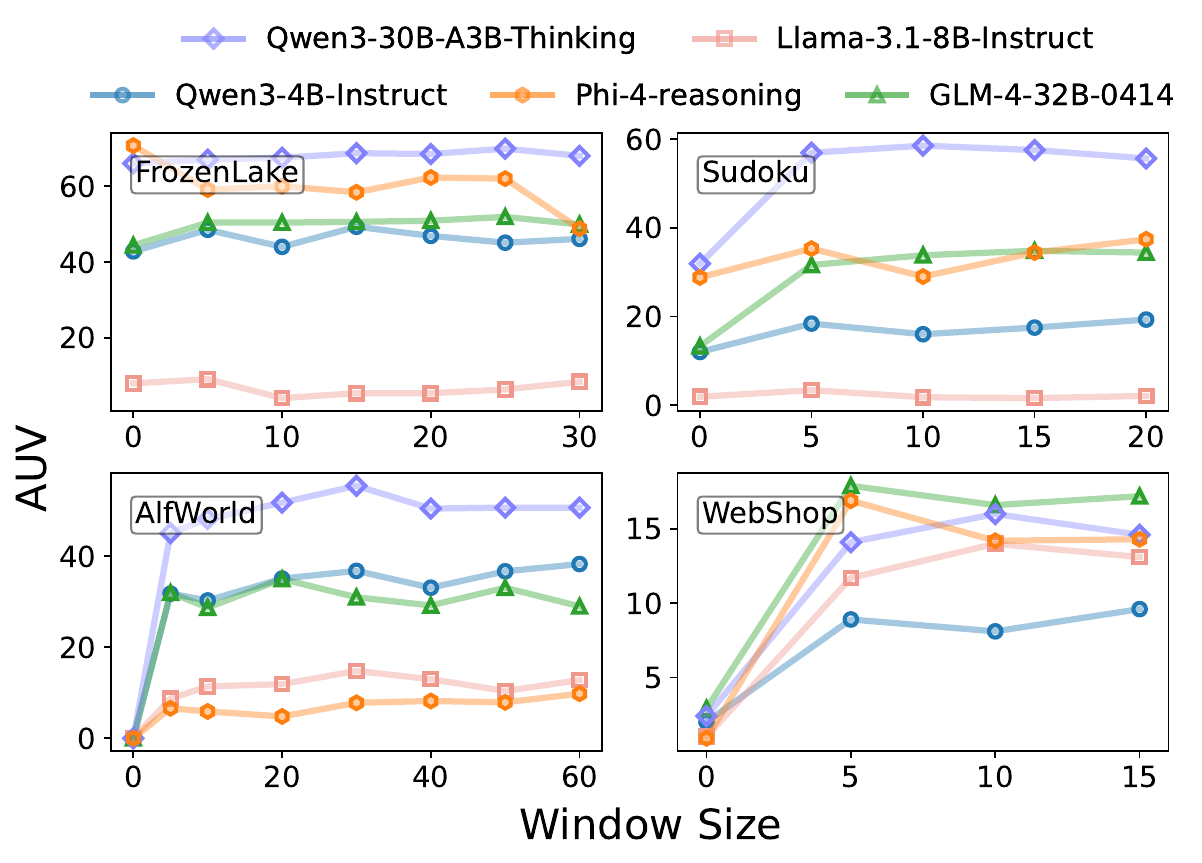}
  \caption{Window size denotes the number of most recent interaction turns retained in memory.}
  \label{fig:win_size}
\end{figure}

\begin{figure*}[t]
  \includegraphics[width=2\columnwidth]{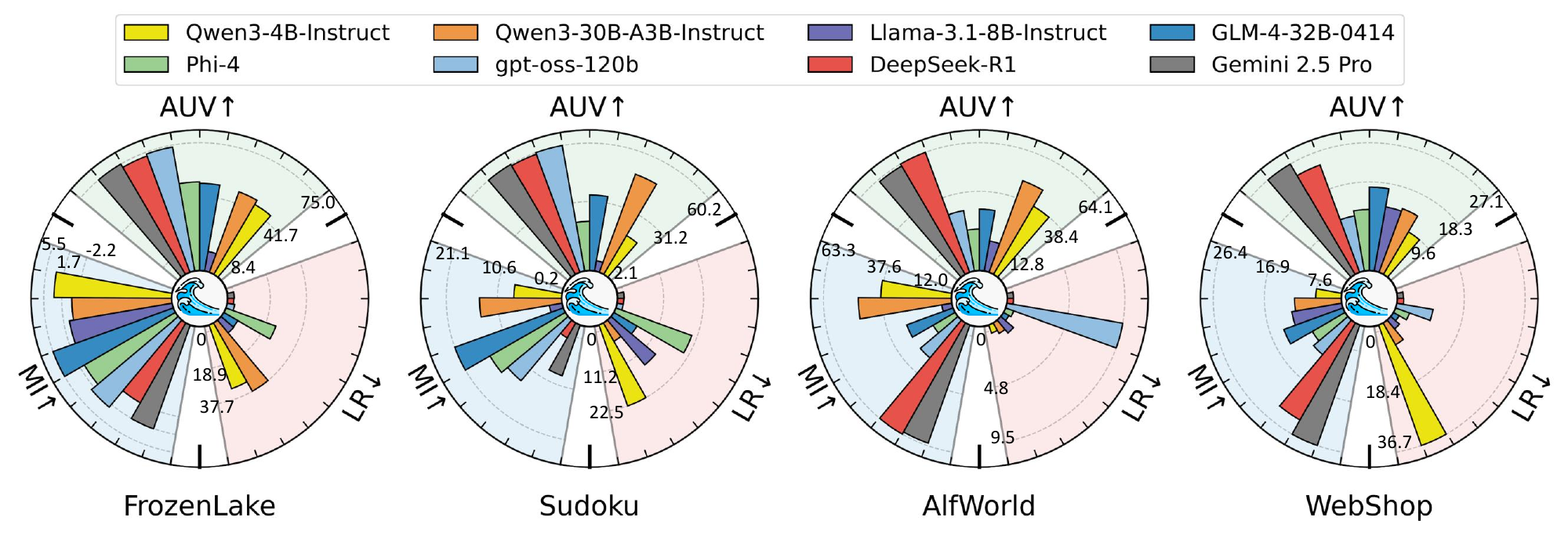}
  \caption{Comprehensive evaluation based on \ours. Bar height stands for the normalized performance on each metric. Uparrow $\uparrow$ indicates that a higher AUV is better. Downarrow $\downarrow$ indicates that a lower LR is better.  Notably, uparrow $\uparrow$ for MI indicates that the agent is sensitive to memory in a specific environment. We present 8 models across 4 environments.  More results can be found in Appendix~\ref{appendix:radar}.}
  \label{fig:radar_chart}
\end{figure*}

\paragraph{Expanding Memory Demonstrates Saturation.}
Given the limited or even negative influence of working memory, we further evaluate the impact of fine-grained working memory length. As illustrated in Figure~\ref{fig:win_size}, we restrict the agent’s access to a limited prefix of recent interaction history. Only several of the most recent interaction turns are visible for the agent at each step, which is denoted as the window size. We can observe that performance gains exhibit diminishing marginal returns. Benefits are confined primarily to the first 5 window sizes, after which the curve rapidly plateaus. This indicates that simply scaling the history buffer yields negligible improvements once the essential context is captured.

\paragraph{Task Structure Determines Memory Efficacy.}
As detailed in Sec~\ref{sec:pre}, we categorize environments into POMDP and MDP frameworks to explain the divergent impact of memory. As illustrated in Figure~\ref{fig:mri}, for partially observable tasks such as WebShop, memory is essential for state reconstruction, leading to positive performance gains. Conversely, in fully observable MDPs like FrozenLake, the environment feedback renders historical context redundant. Models such as \textit{Phi-4-reasoning} often process this redundancy as noise, resulting in a negative influence of working memory. This confirms that memory acts as a necessary resource in information-bound tasks but often functions as a cognitive distraction in pure reasoning domains for models sensitive to redundant history, which provides more insights for future research on working memory management during TTI. More results of MI can be found in Appendix~\ref{appendix:mri}.

\begin{tcolorbox}[
    colback=iceblue,
    colframe=deepblue,
    arc=3mm,
    boxrule=0.8pt,
    left=6pt, right=6pt, top=6pt, bottom=6pt
]
\noindent \textbf{Takeaway 3:} Simply leveraging working memory without further management is often harmful in reasoning-bound scenarios. Even in information-bound scenarios, where the memory is inevitable for task completion, full memory does not guarantee optimality.
\end{tcolorbox}

%% file: Tables/lr.tex
\begin{table}[t]
\centering
\footnotesize
\resizebox{\linewidth}{!}{
\begin{tabular}{l|c|c|c|c|c}
\toprule
\textbf{Model}  &
\textbf{BW} &
\textbf{FL} &
\textbf{Su} &
\textbf{AW} &
\textbf{WS}
\\
\midrule
\multicolumn{6}{c}{\cellcolor{gray!15} Non-thinking Model}  \\
\midrule
\cellcolor{mycellcolor}Qwen3-4B-Instruct & 15.8 & 32.0 & 22.5 & 0.3 & 36.7 \\
\cellcolor{mycellcolor}Qwen3-30B-A3B-Instruct & 1.0 & 37.7 & 4.2 & 0.5 & 5.7 \\
\cellcolor{mycellcolor}Llama-3.1-8B-Instruct & 3.2 & 5.4 & 14.7 & 0.9 & 2.1\\
\cellcolor{mycellcolor}Llama-3.3-70B-Instruct & 0.0 & 9.5 & 2.7 & 0.0 & 1.6 \\
\cellcolor{mycellcolor}Glm-4-9B-Chat & 7.6 & 16.7 & 9.5 & 0.8 & 1.5 \\
\cellcolor{mycellcolor}Glm-4-32B-0414 & 1.2 & 5.1 & 5.8 & 0.0 & 0.1 \\
\cellcolor{mycellcolor}Mistral-7B-Instruct & 51.0 & 63.3 & 15.7 & 10.4 & 17.5 \\
\cellcolor{mycellcolor}Ministral-3-14B-Instruct & 6.0 & 4.3 & 6.9 & 2.7 & 4.0 \\
\cellcolor{mycellcolor}Phi-4 & 12.3 & 24.3 & 21.0 & 0.2 & 2.6 \\
\cellcolor{mycellcolor}DeepSeek-V3.2 & 0.0 & 0.0 & 0.1 & 0.0 & 0.0 \\
\rowcolor{mycellcolor2}Gemini 2.5 Flash & 0.0 & 1.0 & 0.6 & 0.3 & 0.2 \\
\midrule
\multicolumn{6}{c}{\cellcolor{gray!15} Thinking Model}  \\
\midrule
\cellcolor{mycellcolor}Qwen3-4B-Thinking & 28.5 & 9.8 & 34.3 & 4.4 & 4.5 \\
\cellcolor{mycellcolor}Qwen3-30B-A3B-Thinking & 2.0 & 5.3 & 2.2 & 1.3 & 0.7 \\
\cellcolor{mycellcolor}Phi-4-reasoning & 5.2 & 1.4 & 1.0 & 1.4 & 35.7 \\
\cellcolor{mycellcolor}gpt-oss-120b & 1.0 & 0.7 & 0.0 & 9.5 & 9.2 \\
\cellcolor{mycellcolor}DeepSeek-R1 & 0.0 & 0.0 & 0.0 & 0.0 & 0.1 \\
\rowcolor{mycellcolor2}Gemini 2.5 Pro & 0.0 & 0.2 & 0.1 & 0.0 & 0.0 \\
\bottomrule
\end{tabular}
}
\caption{Loop Ratio on five environments: BlocksWorld (BW), FronzenLake (FL), Sudoku (Su), AlfWorld (AW), WebShop (WS). \textcolor[HTML]{289BA2}{Colored model name} represents proprietary models.}
\label{tab:lr}
\end{table}

%% file: Sections/4Framework.tex
\section{Application of TIDE Framework}
\label{sec:framework}



In this section, we synthesize the three metrics to present a holistic TTI analysis of the SOTA LLM agents in the five environments mentioned above.
We further demonstrate the generality of \ours~by applying it in a post-hoc manner to external trajectories, without requiring re-execution of the original experiments.

\subsection{Comprehensive Diagnostic}
\ours~serves as a light-weight tool to deconstruct the underlying mechanisms of TTI into three distinct dimensions.  Figure~\ref{fig:radar_chart} depicts a comparative analysis.
More results across various models can be found in Appendix~\ref{appendix:radar}. We reveal the insights as follows:

\paragraph{Extremely Large-scale Models Stand Out as a Robust Baseline.} \textit{Gemini} and \textit{DeepSeek} series consistently achieve superior AUV and LR across all environments. Notably, in reasoning-bound tasks, these models exhibit relatively low MI. The lower MI does not imply a deficiency in the model's memory utilization capability. This suggests that their robust reasoning capability reduces reliance on interaction memory for optimal performance.

\paragraph{Adaptive Models Derive TTI Capacity from Working Memory With Minimum Loop.}
We identify agents like \textit{GLM-32B-0414} with high memory utility, characterized by a small gap between AUV and SR. This implies their TTI relies almost exclusively on interaction history. Crucially, this success corresponds to a low LR, indicating that memory is effectively merged into behavior adaptation. We highlight that even agents with minimal prior knowledge can attain SOTA performance if they possess sufficient behavior adaption to ground themselves in environmental feedback.

\paragraph{Merely Test-Time Scaling is Insufficient for TTI.} 
We challenge the view that test-time scaling inherently guarantees better behavior adaptation.
Empirically, reasoning-enhanced models, such as \textit{gpt-oss-120b}, often fail to translate internal chain-of-thought into effective external actions in information-bound tasks. This reveals a critical decoupling between internal cognitive capacity and external interactive efficacy. Consequently, agent design should shift from solely maximizing static reasoning depth to explicitly optimizing dynamic, interaction-driven evolution.

\subsection{Generalize to Extended Environments}
\label{sec:gui}
Our proposed framework is deliberately designed to be both agent-agnostic and environment-agnostic, ensuring broad compatibility across diverse evaluation settings. Leveraging this property, we extend \ours~to GUI Agents environments through a secondary analysis of existing interaction logs~\cite{xie2024osworld, bonatti2025windows, liu2025scalecua, rawlesandroidworld, sun2025scienceboard}. 
Notably, \ours~can operate solely on recorded interaction trajectories without re-executing the experiments. Implementation details of the code interface are provided in Appendix~\ref{appendix:code_structure}.

  
As shown in Table~\ref{tab:osworld}, we provide details in OSWorld, separating trajectories based on the presence of loops. We further report the proportion of \textit{Click} actions among all loop actions. Overall, most agent models experience substantial performance degradation on trajectories that contain loops. For instance, \textit{UI-TARS-1.5-72B-DPO} demonstrates an AUV collapse from 26.3 (w/o Loop) to 5.3 (w/ Loop). In contrast, \textit{Claude3.7-Sonnet} exhibits exceptional robustness, maintaining a comparable AUV 6.9 even in scenarios containing loops. This sharp contrast indicates that the high\textit{ Click} action loop ($\sim$50\%) appears to be a major contributor to the performance drop, suggesting that grounding remains the bottleneck of current GUI tasks. 

Moreover, the details in Appendix~\ref{appendix:gui} demonstrate a significant loop phenomenon in both proprietary and open-source models, indicating that loop mitigation is a critical unresolved challenge.

%% file: Sections/5Relatedwork.tex
\section{Related Works}
\input{Tables/osworld}

\paragraph{Multi-turn Interaction Agent.}

While LLMs exhibit strong reasoning capabilities~\citep{deepseekai2025deepseekr1incentivizingreasoningcapability,xu2025phi, yang2025dynamic}, relying purely on internal estimates lacks environmental grounding, especially for real-world environments. Such isolation renders agents prone to hallucination~\citep{du2024haloscope,yan2025mur, huang2025survey} and suboptimal performance in partially observable or non-static scenarios~\citep{parisi2024beyond, wei2025testtimepolicyadaptationenhanced}. Consequently, effective problem-solving requires active self-refinement~\citep{madaan2023self,sun2024corex,cheng2025vision} to adapt action from environmental feedback~\citep{xu2025interactive}. Recent works~\citep{gao2025survey, dou2025evalearn, shen2025thinking, huang2025environmentscalinginteractiveagentic, acikgoz2025selfimprovingllmagentstesttime} are shifting their focus to multi-turn interaction agents, which adopt inner reasoning to dynamic interaction contexts. We formalize these evolving mechanisms as TTI.



\paragraph{LLM Agent Evaluation.} 
Outcome-oriented frameworks~\citep{liu2025agentbenchevaluatingllmsagents, zhou2024webarenarealisticwebenvironment,sun2025genesis} primarily benchmark agents based on the final Success Rate (SR). While effective for overall performance ranking, these metrics treat the task process as a black box, obscuring the interaction costs and efficiency. To provide granular insights, several works have shifted towards fine-grained analysis. AgentBoard~\citep{chang2024agentboard} and AgentQuest~\citep{gioacchini2024agentquest} refine SR into Progress Rate (PR) to quantify partial success. Moreover, subsequent studies~\citep{Jian2024,Chang2025,Yuanzhe2025,Yu2025,zhang2026a3benchbenchmarkingmemorydrivenscientific} have further decomposed performance into specific capabilities, such as planning and tool utilization. However, these methods overlook the temporal dynamics of how an agent evolves and adapts behavior during interaction.



%% file: Tables/osworld.tex
\begin{table}[t]
\centering
\footnotesize
\resizebox{\linewidth}{!}{
\begin{tabular}{l|cc|c}
\toprule
\textbf{Model}  &
\multicolumn{2}{c|}{\textbf{w/ Loop}} &
\multicolumn{1}{c}{\textbf{w/o Loop}}
\\
 & \textbf{AUV} & \textbf{Click Ratio}
 & \textbf{AUV}
\\
\midrule

\rowcolor{mycellcolor}UI-TARS-1.5-7B 
& 3.2 & 57.2
& 28.0 \\

\cellcolor{mycellcolor}UI-TARS-72B-DPO 
& 5.3 & 50.7
& 26.3 \\

\cellcolor{mycellcolor}Qwen2.5-VL-72B-Instruct 
& 0.4 & 50.4
& 8.3 \\

\rowcolor{mycellcolor2}Claude3.7-Sonnet
& 6.9 & 31.1
& 9.0 \\

\rowcolor{mycellcolor2}GPT-4o + ScaleCUA-7B 
& 6.6 & 47.4
& 29.5 \\

\bottomrule
\end{tabular}
}
\caption{Results on OSWorld. Trajectories are divided into two categories based on the presence of loops. Click Ratio is defined as the proportion of \textit{Click} actions among all loop actions. 
\textit{GPT-4o + ScaleCUA-7B} denotes GPT-4o as planner and ScaleCUA as GUI grounder. \textcolor[HTML]{289BA2}{Colored model name} contains proprietary model.}
\label{tab:osworld}
\end{table}

%% file: Sections/6Conclustion.tex
\section{Conclusion}
In this work, we formalize TTI as a dynamic improvement process.
To address the limitations of existing evaluations in multi-turn agent tasks, we introduce \ours, an agent-agnostic and environment-agnostic diagnostic framework that characterizes the dynamics of TTI through three complementary metrics: AUV, LR, and MI.
Through extensive experiments across diverse models and environments, our analysis reveals that agent performance hinges not only on final success but also on how efficiently, adaptively, and effectively agents improve through interaction. We hope this work encourages a shift from evaluating static proficiency toward diagnosing and optimizing agent dynamics.

%% file: Sections/AppendixA.tex
\section{Theoretical Analysis of AUV Properties}
\label{appendix:auv_theory}

In this section, we provide a rigorous theoretical framework to analyze the mathematical properties of the area under variation (AUV) metric. We argue that \textbf{AUV is not merely a redundant correlate of final success rate, but a distinct metric that captures convergence rate and marginal interaction gain}. Moreover, we provide proof of statistical consistency and convergence for AUV.

\subsection{Preliminaries and The Weighted-Increment Lemma}

Let the interaction horizon be defined as $\mathcal{T} = \{t \in \mathbb{Z} \mid 0 \le t < t_{max}\}$.  Let $P_t$ denote the cumulative success rate at turn $t$, which is monotonically non-decreasing. Notably, $P_0=0$. We define the marginal performance gain at step $k$ as $\delta_k$, representing the incremental probability of discovering a solution at the $k$-th interaction turn:
\begin{equation}
    \delta_k = P_{k+1} - P_k, \quad \text{where } \delta_k \ge 0.
\end{equation}
Consequently, the performance at any step $t$ can be reconstructed as:
\begin{equation}
    P_t = \sum_{k=t_{min}}^{t-1} \delta_k
\end{equation}

Recall the definition of AUV, calculated via the discrete trapezoidal integration rule over the normalized horizon $H = t_{max}$:
\begin{equation}
    \label{eq:auv_def_appendix}
    \text{AUV} = \frac{1}{H} \sum_{t=0}^{t_{max}-1}  \frac{P_t + P_{t+1}}{2}
\end{equation}

To facilitate our proofs, we first derive \textbf{Lemma~1} that reformulates AUV as a weighted sum of marginal gains.

\paragraph{Lemma 1: Weighted-Increment Representation}
\textit{The AUV is mathematically equivalent to a linear combination of all marginal gains $\delta_k$, weighted by a time-decaying coefficient $w_k$.}

\textit{Proof.}
First, substitute the recursive expansion $P_t = \sum_{k=t_{min}}^{t-1} \delta_k$ and $P_{t+1} = P_t + \delta_t$ into Eq.~(\ref{eq:auv_def_appendix}). Let $\mathcal{A} = H \cdot \text{AUV}$ denote the area without normalization:
\begin{equation}
    \begin{aligned}
        \mathcal{A} &= \sum_{t=0}^{t_{max}-1} \left( P_t + \frac{1}{2}\delta_t \right) \\
        &= \sum_{t=0}^{t_{max}-1} \left( \sum_{k=0}^{t-1} \delta_k + \frac{1}{2}\delta_t \right)
    \end{aligned}
\end{equation}
We analyze how many times a specific $\delta_k$ generated at step $k$ contributes to the summation over $t$:
\begin{itemize}
    \item The term $\frac{1}{2}\delta_t$ appears exactly once when $t=k$.
    \item The accumulated term $\sum_{k=0}^{t-1} \delta_k$ includes $\delta_k$ for all subsequent time steps $t > k$. Specifically, $\delta_k$ contributes a weight of $1.0$ for each step from $t=k+1$ to $t_{max}-1$.
\end{itemize}
Summing these contributions, the total weight coefficient $w_k$ for $\delta_k$ is:
\begin{equation}
    \begin{aligned}
        w(k) &= \underbrace{0.5}_{\text{contribution at } t=k} + \underbrace{\sum_{t=k+1}^{t_{max}-1} 1}_{\text{contribution at } t > k} \\
        &= 0.5 + ( (t_{max}-1) - (k+1) + 1 ) \\
        &= t_{max} - k - 0.5
    \end{aligned}
\end{equation}
Thus, AUV can be rewritten as follows:
\begin{equation}
    \text{AUV} = \frac{1}{H} \sum_{k=0}^{t_{max}-1} w(k) \delta_k
    \label{eq:lemma1}
\end{equation}
\hfill $\square$

\subsection{Differs from Final Success Rate}

\paragraph{Proposition 1:}
\textit{AUV resolves the ambiguity within the equivalence class of trajectories yielding the same final metric $SR_{\text{final}}$, strictly distinguishing efficient paths from inefficient ones.}

\textit{Proof.}
The final success rate is a path-agnostic metric that performs a compression of the interaction trajectory. In contrast, AUV is a path-dependent metric.

Let the trajectory of marginal gains be represented by a vector $\boldsymbol{\delta} = [\delta_{0}, \dots, \delta_{t_{max}-1}]^T$.
The standard metric $SR_{\text{final}}$ is mathematically equivalent to the unweighted $L_1$ norm shifted by the baseline:
\begin{equation}
    SR_{\text{final}}(\boldsymbol{\delta}) =\|\boldsymbol{\delta}\|_1 = \sum_{k} \delta_k
\end{equation}
This functional is permutation-invariant that any temporal rearrangement of the gain sequence $\{\delta_k\}$ yields the exact same $SR_{\text{final}}$. Thus, $SR_{\text{final}}$ defines an set $\mathcal{C}_{\Omega} = \{ \boldsymbol{\delta} \mid \sum \delta_k = \Omega \}$, where all trajectories achieving total gain $\Omega$ are indistinguishable.

In contrast, AUV is formulated as the inner product:
\begin{equation}
    \text{AUV}(\boldsymbol{\delta}) = \frac{1}{H} \langle \mathbf{w}, \boldsymbol{\delta} \rangle = \frac{1}{H} \sum_{k} w(k) \delta_k
\end{equation}
Since $\mathbf{w}$ is not vector $\mathbf{1}$ that elements are all ones, the value defined by AUV is totally different from the value defined by SR.

To demonstrate the resolving power of AUV, consider a temporal perturbation within the set  $\mathcal{C}_{\Omega}$. 
Assume an agent originally achieves a marginal gain $\epsilon$ at a late step $t_{late}$. We construct a variation where this specific gain is shifted to an earlier step $t_{early}$, forming a new marginal gain trajectory $\boldsymbol{\delta}'$.
So that
$\delta'_{t_{early}} = \delta_{t_{early}} + \epsilon$ and
$\delta'_{t_{late}} = \delta_{t_{late}} - \epsilon$.

This perturbation shows keeps that $ \Delta SR =  0$. And $\Delta \text{AUV} \propto w({t_{early}})\cdot \epsilon - w({t_{late}})\cdot \epsilon  > 0$.

This proves that while $SR_{\text{final}}$ treats performance as a static state variable, AUV treats it as a dynamic process variable. AUV strictly rewards the temporal optimization  of the solution path, retaining structural information lost by the standard metric.

\hfill $\square$

\subsection{Statistical Consistency and Convergence}

In real-world evaluations, the true performance of an agent is an unknown population parameter. We estimate it using a finite dataset of $N$ tasks.
Let $\delta_k^*$ denote the true expected marginal increment $\in \mathbb{R}$ that the agent achieves exactly at step $k$.
The empirical marginal increment observed in the experiment is a random variable, denoted as $\hat{\delta}_k$.

\paragraph{Proposition 2: Consistency and Variance Decay}
\textit{Let $\widehat{\text{AUV}}_N$ be the empirical estimator calculated from $N$ i.i.d. tasks. As $N$ increases, $\widehat{\text{AUV}}_N$ converges in probability to the true population value $\text{AUV}^*$. The estimator is unbiased, and its variance decays at a rate of $O(1/N)$.}

\textit{Proof.}
For a dataset of $N$ tasks, let $x_k^{(i)}$ be the marginal metric increment observed for the $i$-th task at step $k$.
Here, $x_k^{(i)} \in [0, 1]$ represents the value added at this specific step.
The empirical marginal gain $\hat{\delta}_k$ is the sample mean of these individual increments:
\begin{equation}
    \hat{\delta}_k = \frac{1}{N} \sum_{i=1}^{N} x_k^{(i)}
\end{equation}
Consequently, the cumulative performance $P_t$ is the summation of these empirical marginal gains:
\begin{equation}
    P_t = \sum_{j=0}^{t-1} \hat{\delta}_j
\end{equation}

Based on \textbf{Lemma 1}, the empirical AUV is the weighted sum of these marginal gains:
\begin{equation}
    \widehat{\text{AUV}}_N = \frac{1}{H} \sum_{k=0}^{t_{max}-1} w_k \hat{\delta}_k
\end{equation}
Substituting the sample definition of $\hat{\delta}_k$:
\begin{equation}
    \widehat{\text{AUV}}_N = \frac{1}{H} \sum_{k=0}^{t_{max}-1} w_k \left( \frac{1}{N} \sum_{i=1}^{N} x_k^{(i)} \right)
\end{equation}
We rearrange the summation order to isolate individual samples:
\begin{equation}
    \widehat{\text{AUV}}_N = \frac{1}{N} \sum_{i=1}^{N} \underbrace{\left( \frac{1}{H} \sum_{k=0}^{t_{max}-1} w_k x_k^{(i)} \right)}_{Z^{(i)}}
\end{equation}
Here, $Z^{(i)}$ represents a single-task AUV score for the $i$-th task. Since the tasks are i.i.d., the variables $Z^{(1)}, \dots, Z^{(N)}$ are i.i.d. random variables with finite variance $\sigma^2_{Z}$.

Assuming the true expected increment is $\mathbb{E}[x_k^{(i)}] = \delta_k^*$:
\begin{align*}
    \mathbb{E}[\widehat{\text{AUV}}_N] &= \frac{1}{H} \sum_{k} w_k \mathbb{E}[\hat{\delta}_k] \\
    &= \frac{1}{H} \sum_{k} w_k \delta_k^* \\
    &= \text{AUV}^*
\end{align*}

The variance of the sample mean becomes:
\begin{align*}
    \text{Var}(\widehat{\text{AUV}}_N) &= \text{Var}\left( \frac{1}{N} \sum_{i=1}^{N} Z^{(i)} \right)\\
    &= \frac{1}{N^2} \sum_{i=1}^{N} \text{Var}(Z^{(i)}) \\
    &= \frac{\sigma^2_{Z}}{N}
\end{align*}

Since the variance tends to $0$ as $N$ increases, and the estimator is unbiased, $\widehat{\text{AUV}}_N$ converges in probability to $\text{AUV}^*$. This confirms that AUV is a statistically consistent estimator.
\hfill $\square$

%% file: Sections/AppendixB.tex
\section{Supplementary Analysis}
\label{appendix:details}

\subsection{Metrics Necessity Analysis}
\label{necessity}
The proposed metric triad is indispensable for a comprehensive evaluation. AUV is vital for capturing dynamic efficiency. Without it, assessments devolve into static snapshots that obscure convergence rates. LR is critical for differential diagnosis, enabling us to decouple failures rooted in intrinsic reasoning deficits from those stemming from ineffective policy refinement through environment interaction. Finally, MI is essential for causal attribution, allowing us to isolate the precise performance increment attributable solely to memory utilization.

\subsection{Flexible\&Extensible Framework}
\label{appendix:extensible}
Our proposed metric is highly flexible and extensible.
Operationally, metrics such as AUV and LR are derived directly from standard interaction logs, including step-wise success rates and raw trajectory sequences without architectural modifications. Notably, our proposed metric AUV is not only designed for binary success rate. It can also be applied on progress rate or other metics. Furthermore, the computation of MRI necessitates only a straightforward control experiment, restricting the agent's input to the immediate state. Detailed coding help for adopting other logs into our framework can be found in Appendix~\ref{appendix:code_structure}.

\subsection{More Results of SR and AUV}
\label{appendix:auv_validation}
\input{Tables/auv_validation}
To strongly demonstrate that AUV is a new metrics that captures more information than SR, we illustrate some scenarios that a higher SR does not result in a high AUV in Table~\ref{tab:auv_validation}.

\begin{figure}[t]
  \includegraphics[width=\columnwidth]{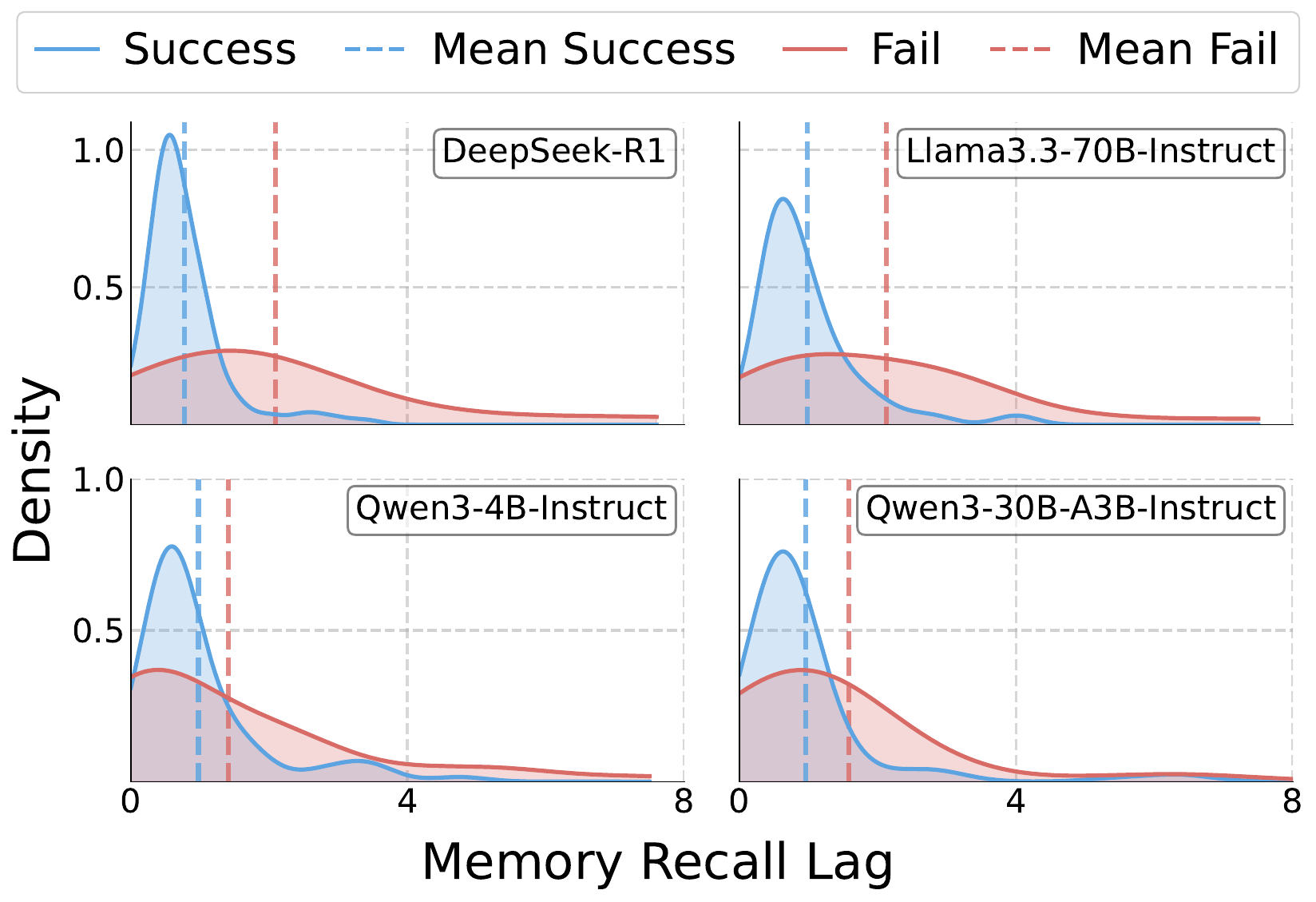}
  \caption{Distribution of memory recall lag in AlfWorld. We separate all trajectories by Success and Fail.}
  \label{fig:recall_distance}
\end{figure}
\input{Tables/summary}
\subsection{Loop Phenomenon Analysis}
\label{appendix:loop_analysis}
\begin{figure*}[t]
  \includegraphics[width=2\columnwidth]{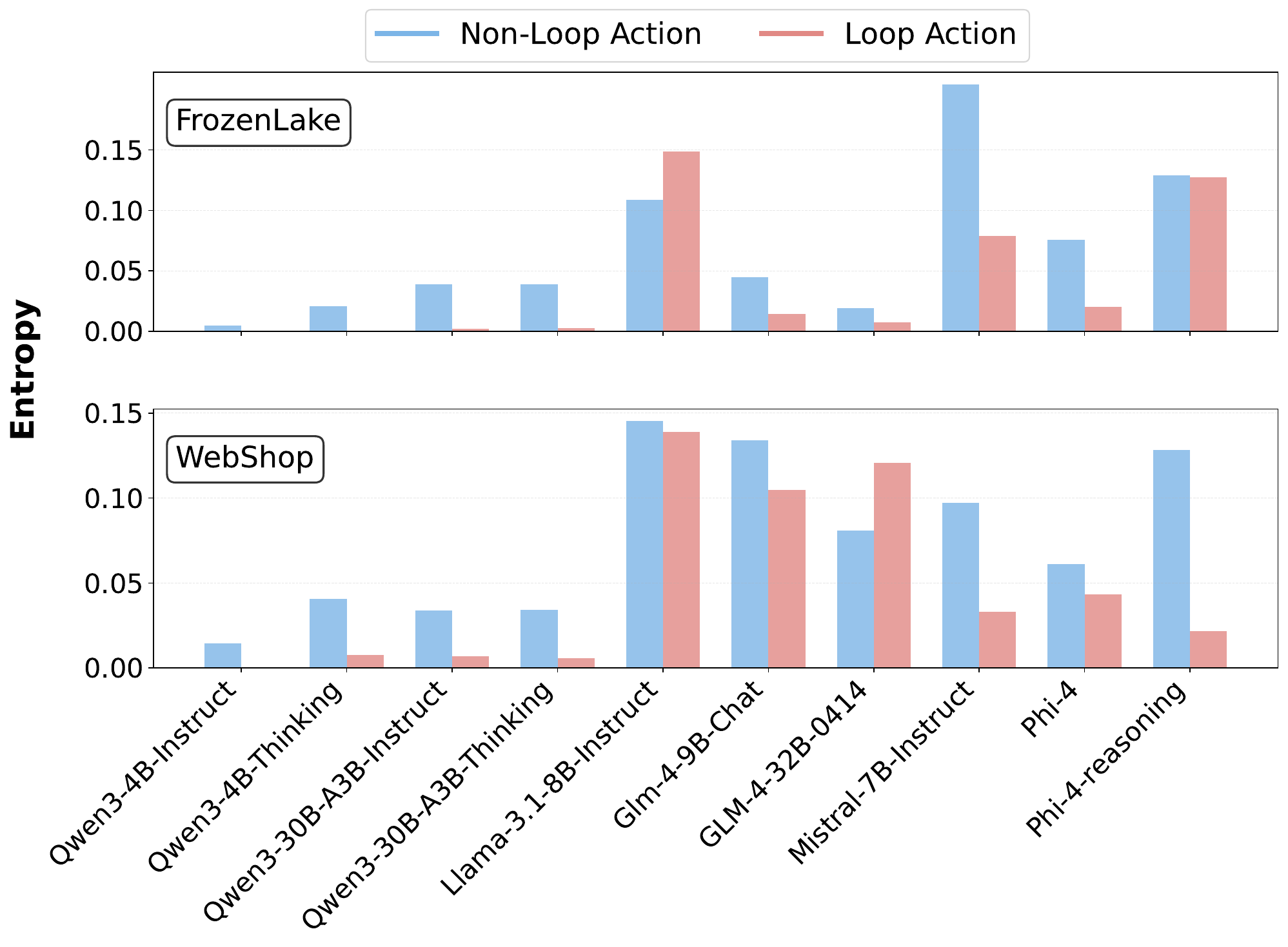}
  \caption{Average action entropy of non-loop actions and loop actions.}
  \label{fig:loop_entropy_bar}
\end{figure*}

To understand the inner cause of loops, we further analyze the action entropy on loop steps against non-loop steps. Figure~\ref{fig:loop_entropy_bar} reveals that most loops stem from over-confidence. The remarkably low entropy during stagnation phases indicates that the agent is confidently fixated on erroneous paths.

Notably, although loops may partially arise from environments with limited action space, our entropy analysis in Appendix~\ref{appendix:loop_analysis} shows that loop actions are significantly associated with over-confidence.
This conclusion is further validated by the results in Sec~\ref{sec:gui}, where agents exhibit significant loops even in GUI environments with high-dimensional and open-ended action spaces.

\subsection{More Results of MI}
\label{appendix:mri}
\input{Tables/mri}
We report detail MI data in Table~\ref{tab:mri}, from which we can observe that a negative MI is widely observed in reasoning-bound tasks.

\input{Tables/gui}
\subsection{Memory Recall Analysis}

To investigate the underlying mechanism of the memory saturation, we define Memory Recall Lag as the temporal interval between acquiring a critical information and its utilization.
In Figure~\ref{fig:recall_distance}, we reveal a sharp divergence in AlfWorld. Successful rollouts cluster tightly around low lags, whereas failures exhibit a long-tail distribution of high lags. This confirms that although the agent retrieves critic information, it fails reasoning across long-range information. Implementation details can be found in Appendix~\ref{appendix:recall_alfworld}.

\subsection{Memory Summary Influence}

To investigate memory utility under alternative management strategies, we instruct the LLM agent to summarize its interaction trajectory and reasoning process. The results in Table~\ref{tab:summary} demonstrate that even with summarization, the accumulated working memory  continues to impose a cognitive burden on LLM agents.

\subsection{More Results of \ours~Comprehensive Analysis}
\label{appendix:radar}
\begin{figure*}[t]
  \includegraphics[width=2\columnwidth]{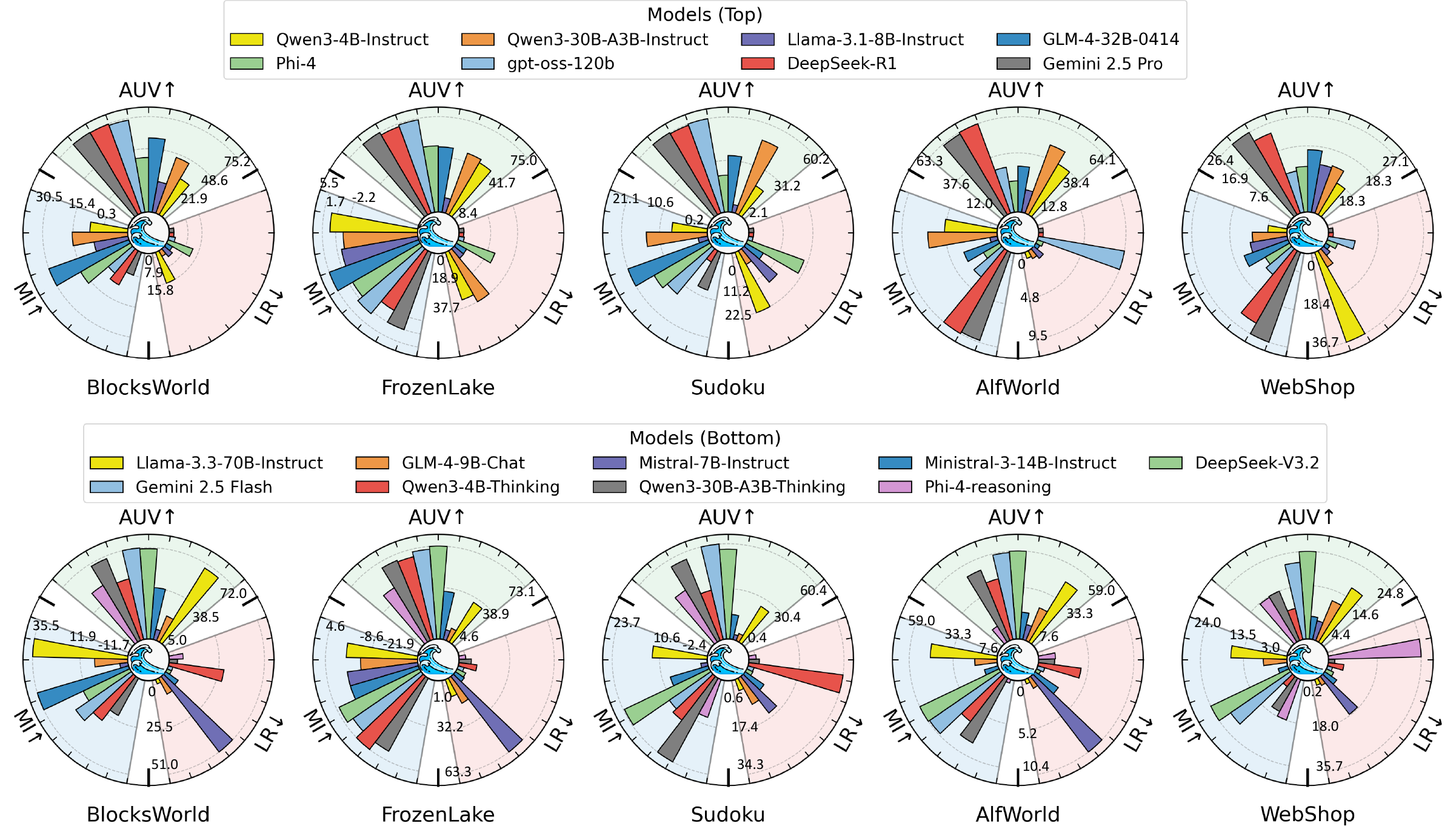}
  \caption{TIDE radar plot results.}
  \label{fig:radar_appendix}
\end{figure*}

We demonstrate all comparative data in Figure~\ref{fig:radar_appendix} to provide a comprehensive diagnose of LLM agents in our paper.

\subsection{More Evaluation Results of Existing GUI Agent Trajectories}
\label{appendix:gui}

As shown in Table~\ref{tab:osworld}, our analysis in OSWorld reveals a stark contrast in robustness. Click Ratio quantifies the proportion of click action loops within all loops. While most agents suffer catastrophic performance degradation in looping trajectories, \textit{Claude3.7-Sonnet-20250219} stands out as the sole exception, maintaining a loop AUV 6.9\% comparable to its non-loop performance 9.0\%. This resilience correlates directly with its significantly lower Click Ratio 31.1\%. Since high click ratios in loops typically indicate repeated failed attempts to manipulate GUI elements, Claude's performance pinpoints grounding precision as the critical bottleneck causing exploration failure in GUI environments.

%% file: Tables/auv_validation.tex
\begin{table*}[t]
\centering
\footnotesize
\resizebox{\linewidth}{!}{
\begin{tabular}{l|l|l|cc|cc}
\toprule
\textbf{Benchmark} &
\textbf{ModelA} &
\textbf{ModelB} &
\textbf{SR$_A$} &
\textbf{SR$_B$} &
\textbf{AUV$_A$} &
\textbf{AUV$_B$} \\
\midrule

\multirow{5}{*}{BlocksWorld}
& Qwen3-4B-Thinking & Phi-4-reasoning & 77.0 & 68.0 & 47.9 & 51.3 \\
& Qwen3-30B-A3B-Thinking & DeepSeek-V3.2 & 98.0 & 98.0 & 69.8 & 71.1 \\
& gpt-oss-120b & DeepSeek-R1 & 100.0 & 100.0 & 73.7 & 74.9 \\
& gpt-oss-120b & Gemini 2.5 Pro & 100.0 & 100.0 & 73.7 & 75.2 \\
& DeepSeek-R1 & Gemini 2.5 Pro & 100.0 & 100.0 & 74.9 & 75.2 \\

\midrule
\multirow{3}{*}{FrozenLake}
& Gemini 2.5 Flash & DeepSeek-V3.2 & 97.0 & 97.0 & 70.7 & 73.1 \\
& Gemini 2.5 Flash & DeepSeek-R1 & 97.0 & 96.0 & 70.7 & 72.6 \\
& Phi-4-reasoning & Qwen3-30B-A3B-Instruct & 62.0 & 61.0 & 48.8 & 49.0 \\

\midrule
\multirow{5}{*}{Sudoku}
& GLM-4-9B-Chat & Llama-3.1-8B-Instruct & 4.0 & 4.0 & 2.0 & 2.1 \\
& GLM-4-32B-0414 & Phi-4-reasoning & 55.0 & 51.0 & 34.4 & 37.4 \\
& Qwen3-4B-Thinking & GLM-4-32B-0414 & 55.0 & 55.0 & 30.7 & 34.4 \\
& Qwen3-4B-Thinking & Phi-4-reasoning & 55.0 & 51.0 & 30.7 & 37.4 \\
& Qwen3-30B-A3B-Thinking & DeepSeek-V3.2 & 95.0 & 93.0 & 55.6 & 56.8 \\

\midrule
\multirow{2}{*}{AlfWorld}
& DeepSeek-V3.2 & Gemini 2.5 Pro & 80.7 & 80.7 & 59.0 & 62.9 \\
& gpt-oss-120b & GLM-4-32B-0414 & 50.7 & 37.9 & 29.0 & 29.0 \\

\midrule
\multirow{10}{*}{WebShop}
& Ministral-3-14B-Instruct & Qwen3-4B-Instruct & 15.2 & 15.0 & 5.4 & 9.6 \\
& Ministral-3-14B-Instruct & Qwen3-4B-Thinking & 15.2 & 12.2 & 5.4 & 8.2 \\
& Llama-3.1-8B-Instruct & Qwen3-30B-A3B-Instruct & 20.8 & 20.8 & 13.1 & 13.7 \\
& Phi-4-reasoning & Qwen3-30B-A3B-Thinking & 22.6 & 21.0 & 14.3 & 14.6 \\
& gpt-oss-120b & Qwen3-30B-A3B-Instruct & 26.2 & 20.8 & 11.1 & 13.7 \\
& Llama-3.3-70B-Instruct & gpt-oss-120b & 26.8 & 26.2 & 17.6 & 11.1 \\
& gpt-oss-120b & GLM-4-9B-Chat & 26.2 & 16.4 & 11.1 & 11.5 \\
& gpt-oss-120b & GLM-4-32B-0414 & 26.2 & 25.4 & 11.1 & 17.2 \\
& gpt-oss-120b & Phi-4 & 26.2 & 18.8 & 11.1 & 12.2 \\
& gpt-oss-120b & Phi-4-reasoning & 26.2 & 22.6 & 11.1 & 14.3 \\

\bottomrule
\end{tabular}
}
\caption{All model pairs $(A, B)$ satisfying $\mathrm{SR}(A) \ge \mathrm{SR}(B)$ and $\mathrm{AUV}(A) \le \mathrm{AUV}(B)$. Three key scenarios of the complementary nature of SR and AUV metrics in model evaluation are illustrated: equal SR with divergent AUV, equal AUV with divergent SR, lower AUV but higher SR, and instances of lower SR but higher AUV between two models.}
\label{tab:auv_validation}
\end{table*}

%% file: Tables/summary.tex
\begin{table*}[t]
\centering
\footnotesize
\resizebox{\linewidth}{!}{
\begin{tabular}{l|cc|cc|cc|cc|cc}
\toprule
\textbf{Model} &
\multicolumn{2}{c|}{\textbf{BlocksWorld}} &
\multicolumn{2}{c|}{\textbf{FrozenLake}} &
\multicolumn{2}{c|}{\textbf{Sudoku}} &
\multicolumn{2}{c|}{\textbf{AlfWorld}} &
\multicolumn{2}{c}{\textbf{WebShop}}    
\\
 & \textbf{Naive} & \textbf{Summary}
 & \textbf{Naive} & \textbf{Summary}
 & \textbf{Naive} & \textbf{Summary}
 & \textbf{Naive} & \textbf{Summary}
 & \textbf{Naive} & \textbf{Summary}
\\
\midrule

Qwen3-4B-Instruct
& 30.8 & 23.5
& 46.1 & 45.7
& 19.3 & 11.9
& 38.3 & 25.5
& 9.6 & 4.8 \\

Llama3.1-8B-Instruct
& 21.9 & 13.5
& 8.4 & 5.3
& 2.1 & 0.7
& 12.8 & 4.4
& 13.1 & 7.9 \\

Glm4-9B-Chat
& 18.2 & 8.1
& 7.2 & 2.7
& 2.0 & 0.0
& 21.8 & 7.6
& 11.5 & 11.6 \\

Mistral-7B-Instruct
& 5.0 & 4.8
& 4.6 & 0.0
& 0.4 & 0.0
& 7.6 & 0.0
& 4.4 & 0.0 \\

\bottomrule
\end{tabular}
}
\caption{Memory summary performance comparison across five environments. We report AUV in two settings: \textbf{Origin} (LLM agents interact with the environment without summary) and  \textbf{Summary} (LLM agents interact with the environment with summary).}
\label{tab:summary}
\end{table*}

%% file: Tables/mri.tex
\begin{table*}[t]
\centering
\resizebox{\linewidth}{!}{
\begin{tabular}{l|cc|cc|cc|cc|cc}
\toprule
\multirow{2}{*}{\textbf{Model}}  &
\multicolumn{2}{c|}{\textbf{BlocksWorld}} &
\multicolumn{2}{c|}{\textbf{FrozenLake}} &
\multicolumn{2}{c|}{\textbf{Sudoku}} &
\multicolumn{2}{c|}{\textbf{AlfWorld}} &
\multicolumn{2}{c}{\textbf{WebShop}} \\
& \multicolumn{1}{c}{\textbf{AUV$\uparrow$}} & \multicolumn{1}{c|}{\textbf{MI}}
& \multicolumn{1}{c}{\textbf{AUV$\uparrow$}} & \multicolumn{1}{c|}{\textbf{MI}}
& \multicolumn{1}{c}{\textbf{AUV$\uparrow$}} & \multicolumn{1}{c|}{\textbf{MI}}
& \multicolumn{1}{c}{\textbf{AUV$\uparrow$}} & \multicolumn{1}{c|}{\textbf{MI}}
& \multicolumn{1}{c}{\textbf{AUV$\uparrow$}} & \multicolumn{1}{c}{\textbf{MI}} \\
\midrule
\multicolumn{11}{c}{\cellcolor{gray!15} Non-thinking Model}  \\
\midrule
\cellcolor{mycellcolor}Qwen3-4B-Instruct 
& 30.8 & 7.0 
& 46.1 & 3.3 
& 19.3 & 7.3 
& 38.3 & 38.3 
& 9.6  & 7.6  \\
\cellcolor{mycellcolor}Qwen3-30B-A3B-Instruct
& 45.4 & 15.8 
& 49.0 & -0.5 
& 48.8 & 14.3 
& 48.1 & 48.1 
& 13.7 & 11.4  \\
\cellcolor{mycellcolor}Llama-3.1-8B-Instruct 
& 21.9 & 5.4 
& 8.4  & 0.4 
& 2.1  & 0.2 
& 12.8 & 12.7 
& 13.1 & 12.1  \\
\cellcolor{mycellcolor}Llama-3.3-70B-Instruct
& 68.2 & 35.5 
& 34.7 & -1.4 
& 24.9 & 12.7 
& 46.7 & 46.7 
& 17.6 & 16.6  \\
\cellcolor{mycellcolor}Glm-4-9B-Chat 
& 18.2 & 4.7 
& 7.2  & -5.5 
& 2.0  & -2.4 
& 21.8 & 20.6
& 11.5 & 8.7  \\
\cellcolor{mycellcolor}Glm-4-32B-0414 
& 57.5 & 30.5 
& 49.9 & 5.5 
& 34.4 & 21.1 
& 29.0 & 29.0 
& 17.2 & 14.3  \\
\cellcolor{mycellcolor}Mistral-7B-Instruct
& 5.0  & -7.7 
& 4.6  & -1.5 
& 0.4  & -0.6 
& 7.6  & 7.6 
& 4.4  & 3.0  \\
\cellcolor{mycellcolor}Mistral-3-14B-Instruct
& 39.0 & 35.5 
& 35.8 & -1.5 
& 13.3 & 8.4 
& 15.8 & 15.8 
& 5.4  & 5.4  \\
\cellcolor{mycellcolor}Phi-4
& 41.2 & 16.8 
& 50.8 & 0.6 
& 21.4 & 15.3 
& 18.6 & 18.6 
& 12.2 & 9.8  \\
\cellcolor{mycellcolor}DeepSeek-V3.2 
& 71.1 & 13.8 
& 73.1 & 3.8
& 56.8 & 23.4 
& 59.0 & 59.0
& 24.8 & 24.0  \\
\rowcolor{mycellcolor2}
Gemini 2.5 Flash 
& 72.0 & 21.7 
& 70.7 & 1.9 
& 60.4 & 11.4 
& 58.1 & 58.1 
& 21.6 & 20.6  \\
\midrule
\multicolumn{11}{c}{\cellcolor{gray!15} Thinking Model}  \\
\midrule
\cellcolor{mycellcolor}Qwen3-4B-Thinking 
& 47.9 & 15.8 
& 66.7 & 4.6 
& 30.7 & 12.7 
& 40.9 & 40.9 
& 8.2  & 6.1  \\
\cellcolor{mycellcolor}Qwen3-30B-A3B-Thinking 
& 69.8 & 9.3 
& 68.0 & 2.0 
& 55.6 & 23.7 
& 50.7 & 50.7 
& 14.6 & 12.2  \\
\cellcolor{mycellcolor}Phi-4-reasoning
& 51.3 & -11.7 
& 48.8 & -21.9 
& 37.4 & 8.6 
& 9.8  & 9.8 
& 14.3 & 13.4 \\
\cellcolor{mycellcolor}gpt-oss-120b
& 73.7 & 4.0 
& 73.8 & 2.4 
& 60.2 & 13.7 
& 29.0 & 29.0 
& 11.1 & 11.1  \\
\cellcolor{mycellcolor}DeepSeek-R1 
& 74.9 & 7.7 
& 72.6 & -2.2 
& 59.4 & 1.1 
& 64.1 & 63.3 
& 24.1 & 23.2  \\
\rowcolor{mycellcolor2}
Gemini 2.5 Pro 
& 75.2 & 0.3 
& 75.0 & 1.7 
& 60.2 & 8.6 
& 62.9 & 62.2 
& 27.1 & 26.1  \\
\bottomrule
\end{tabular}
}
\caption{Detailed MI data. \textcolor[HTML]{289BA2}{Colored model name} represents proprietary models.}
\label{tab:mri}
\end{table*}

%% file: Tables/gui.tex
\begin{table}[t]
\centering
\footnotesize
\resizebox{\linewidth}{!}{
\begin{tabular}{l|cc|cc|cc}
\toprule
\textbf{Model}  &
\multicolumn{2}{c|}{\textbf{AW}} &
\multicolumn{2}{c|}{\textbf{OSWorld}} &
\multicolumn{2}{c}{\textbf{WAA}}
\\
 & \textbf{AUV} & \textbf{LR} 
 & \textbf{AUV} & \textbf{LR}
 & \textbf{AUV} & \textbf{LR}
\\
\midrule
\cellcolor{mycellcolor}InternVL3.5-30B-A3B & 12.4 & 6.8 & -- & -- & 6.7 & 4.4 \\
\rowcolor{mycellcolor}UI-TARS-1.5-7B & 25.0 & 44.8 & 15.4 & 21.3 & -- & -- \\
\cellcolor{mycellcolor}UI-TARS-72B-DPO & 34.8 & 16.9 & 20.3 & 16.1 & -- & -- \\
\cellcolor{mycellcolor}Qwen2.5-VL-72B-Instruct & -- & -- & 5.9 & 4.6 & 6.0 & 8.1 \\
\cellcolor{mycellcolor}Qwen3-VL-235B-A22B-Instruct & 43.2 & 32.6 & -- & -- & 9.5 & 19.1\\

\rowcolor{mycellcolor2}Claude3.7-Sonnet-20250219 & -- & -- & 8.4 & 1.4 & 0.0 & 21.6 \\
\rowcolor{mycellcolor2}GPT-4o + ScaleCUA-7B  & -- & -- & 15.8 & 36.8 & -- & -- \\
\bottomrule
\end{tabular}
}
\caption{Evaluation results in AndroidWorld (AW), OSWorld, WindowsAgentArena (WAA). For \textit{GPT-4o + ScaleCUA-7B}, the former acts as a planner for high-level planning and the latter acts as a grounder for low-level execution. \textcolor[HTML]{289BA2}{Colored model name} contains proprietary model.}
\label{tab:gui}
\end{table}

%% file: Sections/AppendixC.tex
\section{Implementation Details}

\subsection{Experiment Environment}
\label{appendix:environments}
We select 5 widely used environments, dividing them into two categories based on the main character of environment feedback in TTI: (1) reasoning-bound (MDP), including BlocksWorld, FrozenLake, Sudoku. (2) information-bound (POMDP), including AlfWorld, WebShop.

Additionally, we apply our framework to 3 domains of existing POMDP agent interaction trajectories in AndroidWorld, OSWorld and WindowsAgentArena.

\textbf{BlocksWorld} is a PDDL~\citep{vallati20152014} environment in which an agent must rearrange stacks of blocks to satisfy goal configurations. In our experiments, it contains 100 data generated by ourselves, which are included in our data file.

\textbf{FrozenLake} is a grid-world environment in which an agent must navigate toward a goal while avoiding holes. We follow the implementation in RAGEN~\citep{wang2025ragen}. In our experiments, it contains 100 data generated by ourselves, which are included in our data file.

\textbf{Sudoku} is a constraint-satisfaction puzzle in which agent must fill a partially specified grid so that each row, column, and designated subregion contains all required symbols exactly once. In our experiments, it contains 100 data generated by ourselves, which are included in our data file.

\textbf{AlfWorld}~\citep{shridhar2021alfworld} consists of household manipulation tasks that require agents to navigate their environment and execute actions grounded in commonsense reasoning. We use evaluation in distribution, which contains 140 data entries.

\textbf{WebShop}~\citep{yao2022webshop} serves as a network-based simulation environment that enables controlled experiments on e-commerce tasks. We use the full 500 data entries.

\textbf{AndroidWorld}~\citep{rawlesandroidworld} is a real-world Android environment and benchmark designed to evaluate autonomous AI agents on their ability to execute complex, cross-application daily tasks on a mobile operating system.

\textbf{OSWorld}~\citep{xie2024osworld} is a benchmark and realistic computer environment designed to evaluate multimodal agents on their ability to execute open-ended, cross-application tasks within real desktop operating systems.

\textbf{WindowsAgentArena}~\citep{bonatti2025windows} is a scalable benchmark and realistic environment developed by Microsoft to evaluate multimodal AI agents on their ability to execute diverse tasks within a native Windows operating system.

\subsection{Model Selection}
Following recent work~\citep{dou2025evalearn}, we divide models into thinking and non-thinking models. Additionally, we divide models into proprietary models and open-source models. For \textit{Qwen} series, we use a single base model and change between instruct mode and thinking mode, which can be simply implemented by a single parameter \textit{$enable\_thinking$} in function \textit{$apply\_chat\_template$}.

\subsection{Loop Ratio Implementation Details}
\label{appendix:lr}
We provide detailed pseudo code of LR calculation in Algorithm~\ref{alg:loop_ratio_global}. Moreover, for BlocksWorld, FrozenLake, Sudoku, AlfWorld and WebShop, we use exact text match to represent the state and the action. For GUI environments, we use \textit{clip-vit-base-patch32}~\citep{radford2021learning} model for picture embedding and  set  0.999 as the threshold for identifying two pictures as the same state. We use exact match for same action identification. 

\input{Tables/algorithm}

\subsection{$t_{max}$ Configuration}
\label{appendix:t_min_max}

\begin{figure*}[t]
  \includegraphics[width=2\columnwidth]{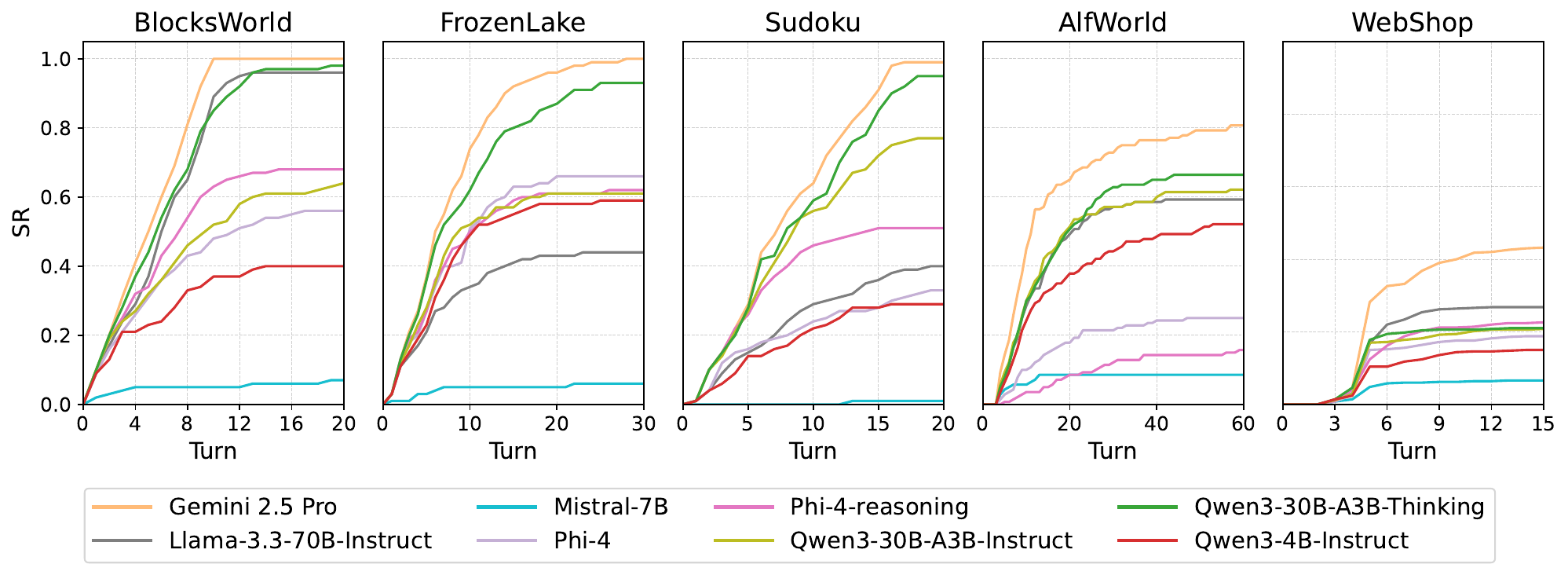}
  \caption{Cumulative success rate curves across five environments. The trajectories illustrate the performance saturation points, which serve as the empirical basis for determining the upper boundary $t_{max}$ in our AUV metric.}
  \label{fig:tminmax}
\end{figure*}

The AUV metric is specifically designed to quantify the efficacy of agent environment interaction. Regarding the upper boundary $t_{max}$, we determine it empirically based on the performance saturation point observed across agents. 

Guided by the cumulative success rate trajectories in Figure~\ref{fig:tminmax}, we set domain-specific $t_{max}$ values of 20, 30, 20, 60, and 15 for BlocksWorld, FrozenLake, Sudoku, AlfWorld, and WebShop, respectively.
Additionally, we set $t_{max}$ to 50 for GUI tasks following the implementation of original data.

Notably, in environments that involve multi-step tasks which cannot be solved within a single action, the AUV metric is inherently bounded away from 1. Nevertheless, an empirical upper bound of AUV can be estimated through extensive rollouts of SOTA models under the same evaluation protocol. An observed AUV value substantially lower than this empirical upper bound indicates untapped potential for improving TTI capacity, rather than a limitation imposed by the task structure itself.


\subsection{Memory Recall Distance in AlfWorld}
\label{appendix:recall_alfworld}
To quantitatively evaluate an agent's ability to maintain and utilize memory of object locations across extended action sequences, we propose the Memory Recall Distance metric. This metric measures the temporal gap between observing a task-relevant object and later interacting with it.

To compute the memory recall distance in a trajectory $\tau$ with $T$ timesteps, we first identify the set of task-relevant objects $\mathcal{G}_{\text{task}}$ by extracting target required objects from the instruction based on the task categorization defined in the AlfWorld benchmark~\citep{shridhar2021alfworld}:
$\mathcal{G}_{\text{task}} = \{ obj \mid obj \text{ is required for task} \}$

Throughout the trajectory, we track when each object appears in the agent's observations. Let $\mathcal{E}(o_t)$ denote the extraction function that returns all objects visible in observation $o_t$. For any object $obj$, we maintain a record of its most recent observation timestep prior to $t$:
\begin{equation}
    LastSeen(obj, t) = \max \{ k\mid k<t, \ obj \in \mathcal{E}(o_k) \}
\end{equation}

where $LastSeen(obj, t)$ is undefined if $obj$ has not been observed before timestep $t$.

The key measurement occurs when the agent interacts with a task-relevant object. Specifically, at timestep $t$, if action $a_t$ involves interaction with object $obj$ where $obj \in \mathcal{G}_{\text{task}}$ and $LastSeen(obj, t)$ is defined, we compute the recall distance as:
$d_t(obj) = t - LastSeen(obj, t)$

This distance captures how many timesteps have passed since the agent last saw this object before deciding to interact with it. Finally, we aggregate these individual recall distances across the entire trajectory to obtain the overall memory recall distance:
\begin{equation}
   D_{\text{recall}}(\tau) = \frac{1}{|\mathcal{I}|} \sum_{(t, obj) \in \mathcal{I}} d_t(obj) 
\end{equation}

where $\mathcal{I}$ represents the set of all valid (timestep, object) pairs in the trajectory:
$\mathcal{I} = \{(t, obj) \mid obj \in \mathcal{G}_{\text{task}} \wedge \exists LastSeen(obj, t)\}$

\subsection{Radar Plot Details}
Specifically, we generate distinct radar profiles for each task environment. To facilitate cross-model comparison, all metrics are normalized to the unit interval [0,1] via min-max scaling across all models. Notably, regarding that LR serves as a negative indicator of TTI performance, we apply an inversion transformation 1 - LR to ensure a consistent orientation across all axes, so that a lager value indicates a better TTI. More over, we add a minimum baseline value to make each bar visible even its corresponding indicator is zero. We also limit the maximum hight so that the highest bar does not reach the boundary of the figure. Notably, these shifts do not influence the diagnose of each model.

\subsection{Framework Design}
\label{appendix:code_structure}
Our framework adopts a highly modular architecture with clear separation of concerns. Complex concurrency control and inference logic are encapsulated within abstract base classes, allowing researchers to focus solely on task-specific adaptations. This design significantly reduces the barrier to entry for incorporating new evaluation benchmarks.

\textbf{Task Execution} The \texttt{TaskRunner} module implements a template method pattern providing flexible lifecycle management while minimizing the required interface surface. Researchers can leverage two complementary environment management strategies: (1) \emph{Resource pooling} for stateful, expensive-to-initialize environments (e.g., WebShop) through the \texttt{EnvPool} mechanism, eliminating redundant instantiation overhead; (2) \emph{On-demand instantiation} for environments requiring complete teardown between episodes, where researchers simply define construction logic in \texttt{\_initialize\_trajectory} and the framework handles per-trajectory environment generation automatically.

\textbf{Context Management} The \texttt{ContextManager} module achieves complete separation between prompt engineering and control flow logic. Interaction histories are abstracted as structured \texttt{StepMemory} objects, enabling seamless switching between experimental configurations (e.g., inclusion of chain-of-thought reasoning, history truncation) without modifying underlying implementation code.

\textbf{Unified Agent Interface} To support fair comparison between open-weight models and proprietary APIs, we provide a unified \texttt{BaseAgent} interface with \texttt{ClientAgent} implementation wrapping OpenAI-format API calls. This abstraction allows seamless backend switching through configuration changes alone---transitioning from local \textit{Qwen3-4B-Instruct} to remote \textit{Gemini 2.5 Pro} requires no task code modifications. The framework incorporates thread-pooled concurrent requests with backoff retry mechanisms, ensuring robust performance despite network latency.

\textbf{Parallel Execution} The framework implements multi-level parallelism for large-scale evaluation on limited computational resources. At the GPU level, automatic dataset sharding distributes workloads across devices with file-lock mechanisms ensuring safe concurrent result aggregation. At the batch level, \texttt{TaskRunner} maintains a global trajectory pool, automatically collecting contexts from all active trajectories into unified batches for efficient inference. 

Furthermore, we present the primary class structures and function interfaces below, covering: core data structures, task runner framework, agent interface, context manager, and environment interface. Notably, we are thankful to valuable works~\cite{wang2025ragen, xue2025simpletir} for their open-sourced code.

\input{CoreCode/datastructure}
\input{CoreCode/taskrunner}
\input{CoreCode/agent}
\input{CoreCode/ctxmanager}
\input{CoreCode/environment}

\subsection{Hyper-Parameter and Hardware Settings}
We set temperature to $0.7$ and top-$p = 1.0$ for all generation. For models with configurable reasoning effort (\textit{gpt-oss-120b}, \textit{Gemini 2.5 Pro}), we use the \textit{medium} setting.
We conduct our experiments on Nvidia-A100 GPUs.

\subsection{Prompt Details}
We provide our detailed prompt implementation. First, we provide the template of interaction format. Detailed system prompt can also be found following the interaction template.

\input{Prompt/interaction_template}

\input{Prompt/blocksworld}

\input{Prompt/frozenlake}

\input{Prompt/sudoku}

\input{Prompt/alfworld}

\input{Prompt/webshop}

%% file: Tables/algorithm.tex
\begin{algorithm*}[t]
\caption{Finite State Machine-based Loop Ratio (LR) Calculation}
\label{alg:loop_ratio_global}

\renewcommand{\algorithmicrequire}{\textbf{Input:}}
\renewcommand{\algorithmicensure}{\textbf{Output:}}

\begin{algorithmic}[1]
\REQUIRE Trajectories set $\{\tau^{(1)}, \tau^{(2)}, \dots, \tau^{(N)}\}$, where each $\tau^{(k)}$ is a trajectory.
\ENSURE LR

\STATE \textbf{Initialize Global Counters:} 
\STATE \quad $S_{\text{total}} \leftarrow 0$ \text{// Total interaction steps across all samples}
\STATE \quad $L_{\text{total}} \leftarrow 0$ \text{// Total loop  steps across all samples}

\FOR{$k = 1$ \textbf{to} $N$}
    \STATE $\tau \leftarrow \tau^{(k)}$
    \STATE $S_{\text{total}} \leftarrow S_{\text{total}} + \text{Length}(\tau)$
    
    \STATE \text{// Reset local tracking variables for the new trajectory}
    \STATE $\mathcal{H} \leftarrow \emptyset$, $l_{\text{prev}} \leftarrow \emptyset$, $t_{\text{end}} \leftarrow -1$

    \FOR{$t = 0$ \textbf{to} \text{Length}($\tau)$}
        \STATE $h_t \leftarrow \text{Hash}(s_t)$ \text{// Compute hash of the current state}
        
        \IF{$h_t \in \mathcal{H}$}
            \STATE $i \leftarrow \mathcal{H}[h_t]$ \text{// Retrieve start index of potential cycle}
            \STATE $l_{\text{curr}} \leftarrow \tau[i : t]$ 
            
            \STATE \text{// Condition 1: Check Non-Recursive}
            \IF{\textbf{not} $\text{HasNested}(l_{\text{curr}})$} 
                
                \STATE \text{// Condition 2: Check Consecutive Repetition}
                \IF{$(i == t_{\text{end}})$ \textbf{and} $(l_{\text{curr}} == l_{\text{prev}})$}
                    \STATE $L_{\text{total}} \leftarrow L_{\text{total}} + (t - i)$ \text{// Accumulate loop length into global counter}
                \ENDIF
                
                \STATE $l_{\text{prev}} \leftarrow l_{\text{curr}}$
                \STATE $t_{\text{end}} \leftarrow t$
            \ENDIF
            
            \STATE $\mathcal{H}[h_t] \leftarrow t$ \text{// Update the latest visit time for this state}
        \ELSE
            \STATE $\mathcal{H}[h_t] \leftarrow t$ \text{// Record the first visit time}
        \ENDIF
    \ENDFOR
\ENDFOR

\STATE \textbf{Return} $\text{LR} = L_{\text{total}} / S_{\text{total}}$
\end{algorithmic}
\end{algorithm*}

%% file: CoreCode/datastructure.tex
\begin{figure*}
\begin{lstlisting}[language=Python, caption={Core Data Structures}, label={lst:data_structures}]
@dataclass
class StepMemory:
    """Stores information for a single reasoning step in the trajectory.
    
    Attributes:
        observation: Current environment observation
        true_state: Ground truth state from environment
        input_state: State representation input to the model
        analysis: Model's reasoning/thinking for this step
        action: Action taken by the model
        is_valid: Whether the action was valid
        feedback: Environment feedback message
        previous_memory: Link to previous step
    """
    observation: str
    true_state: str
    input_state: str
    analysis: str
    action: str
    is_valid: bool
    feedback: str
    previous_memory: Optional['StepMemory']

@dataclass
class TrajectoryInfo:
    """Encapsulates complete trajectory information for parallel processing.
    
    Attributes:
        idx_in_batch: Index within the current batch
        traj_rollout_idx: Rollout index for multiple trajectories
        env: Environment instance for this trajectory
        env_idx: Index in the environment pool
        ctx_manager: Context manager for prompt formatting
        steps: List of step dictionaries
        done: Whether trajectory is completed
        success: Whether task was successfully solved
        stop_right: Whether agent stopped at correct time
    """
    idx_in_batch: int
    traj_rollout_idx: int
    env: BaseEnv
    env_idx: int
    ctx_manager: ContextManager
    steps: List[Dict[str, Any]]
    done: bool
    success: bool
\end{lstlisting}
\end{figure*}

%% file: CoreCode/taskrunner.tex
\begin{figure*}
\begin{lstlisting}[language=Python, caption={Task Runner Framework}, label={lst:task_runner}]
class TaskRunnerBase:
    """Base class for task execution and evaluation management."""
    
    def __init__(self, config: DictConfig, agent: BaseAgent, 
                 dataset: List[Dict]) -> None:
        """Initialize task runner with configuration, agent, and dataset.
        
        Args:
            config: Configuration object containing task and agent settings
            agent: Agent instance for generating actions
            dataset: List of task instances to evaluate
        """
        
    def _prepare_step_history(self, traj: TrajectoryInfo) -> None:
        """Prepare history and state information for current step.
        
        Args:
            traj: Trajectory to prepare step for
        """
        
    def _process_step(self, traj: TrajectoryInfo, 
                      step_info: Dict[str, Any]) -> None:
        """Execute action in environment and update trajectory state.
        
        Args:
            traj: Trajectory to process
            step_info: Dictionary containing action and analysis
        """
        
    def _initialize_trajectory(self, data_idx: int, data: dict, 
                              traj_rollout_idx: int) -> TrajectoryInfo:
        """Initialize single trajectory with environment and context.
        
        Args:
            data_idx: Index in dataset
            data: Data instance dictionary
            traj_rollout_idx: Rollout index for this trajectory
            
        Returns:
            TrajectoryInfo: Initialized trajectory object
        """
        
    def run(self, dp_idx: int = 0, 
            lock: Optional[Lock] = None) -> None:
        """Main entry point for running evaluation on dataset.
        
        Args:
            dp_idx: Starting index in dataset
            lock: Optional lock for thread synchronization
        """
        
    def _run_stepwise_episode_batch(self, 
                                    batch: List[Dict]) -> List[Dict]:
        """Run batch of episodes with step-by-step execution.
        
        Args:
            batch: List of data instances
            
        Returns:
            List of trajectory results with multiple rollouts per instance
        """
\end{lstlisting}
\end{figure*}

%% file: CoreCode/agent.tex
\begin{figure*}
\begin{lstlisting}[language=Python, caption={Agent Interface}, label={lst:agent_interface}]
class BaseAgent:
    """Abstract base class for all agent implementations."""
    def __init__(self, config: DictConfig):
        """Initializes the agent with task-specific configurations.
        
        Args:
            config: Configuration object containing model and generation parameters.
        """
    
    def get_next_step_parallel(self, 
                               trajectories: List[TrajectoryInfo]
                               ) -> List[Dict[str, Any]]:
        """Generate actions for multiple trajectories in parallel.
        
        Args:
            trajectories: List of trajectories to process
            
        Returns:
            List of action dictionaries for each trajectory
        """
        
    def close(self) -> None:
        """Clean up agent resources and connections."""

\end{lstlisting}
\end{figure*}

%% file: CoreCode/ctxmanager.tex
\begin{figure*}
\begin{lstlisting}[language=Python, caption={Context Manager}, label={lst:ctx_manager}]

class ContextManager:
    """Manages conversation history and prompt formatting."""
    
    def __init__(self, system_prompt: str, instruction_prompt: str,
                 tokenizer: AutoTokenizer, config: DictConfig) -> None:
        """Initialize with prompts and formatting configuration.
        
        Args:
            system_prompt: System-level instruction prompt
            instruction_prompt: Task-specific instruction
            tokenizer: Tokenizer for applying chat templates
            config: Configuration for formatting options
        """
        
    def format_prompt(self) -> str:
        """Format full prompt string based on chat format and history.
        
        Returns:
            Formatted prompt string including conversation history
            
        Supported formats:
            - 'default_format': Standard prompt format
            - 'user_assistant_format': Dialogue-based format
            - 'user_assistant_format_part': Partial dialogue format
        """
        
    def format_messages(self) -> List[Dict[str, str]]:
        """Format as message list for API-based agents.
        
        Returns:
            List of message dictionaries with 'role' and 'content' keys
        """
\end{lstlisting}
\end{figure*}

%% file: CoreCode/environment.tex
\begin{figure*}
\begin{lstlisting}[language=Python, caption={Environment Interface}, label={lst:env_interface}]
class BaseEnv:
    """Abstract base class for task environments."""
    
    def reset(self, seed: int, mode: str) -> None:
        """Reset environment to initial state.
        
        Args:
            seed: Random seed for reproducibility
            mode: Evaluation mode ('easy', 'medium', 'hard')
        """
        
    def step(self, action: str) -> Tuple[Any, float, bool, Dict]:
        """Execute action and return environment response.
        
        Args:
            action: Action string to execute
            
        Returns:
            Tuple containing:
                - observation: New environment observation
                - reward: Reward signal
                - done: Whether episode is terminated
                - info: Additional information dictionary
        """
        
    def render(self) -> str:
        """Generate string representation of current state.
        
        Returns:
            Human-readable string of environment state
        """
        
    @property
    def instruction_text(self) -> str:
        """Return task instruction or query text.
        
        Returns:
            Task-specific instruction string
        """
\end{lstlisting}
\end{figure*}

%% file: Prompt/interaction_template.tex
\begin{figure*}[t]
\begin{tcolorbox}[title={Interaction Template}]

System Prompt: [System Prompt] + [Fewshot Trajectory] 

\tcblower 

\begin{tcolorbox}[colback=blue!5, colframe=blue!30, boxrule=0.5pt, 
                  left=2mm, right=2mm, top=1mm, bottom=1mm]
\textbf{User:} [Task Instruction] + \texttt{<state>}[Observation of Turn 1]\texttt{</state>}

\textbf{Assistant:} \texttt{<analysis>}[Agent Reasoning Process at Turn 1]\texttt{</analysis>} 

\texttt{<action>}[Agent Action at Turn 1]\texttt{</action>}
\end{tcolorbox}

\vspace{0.3em}

\begin{tcolorbox}[colback=green!5, colframe=green!30, boxrule=0.5pt,
                  left=2mm, right=2mm, top=1mm, bottom=1mm]
\textbf{User:} \texttt{<state>}[Observation of Turn 2]\texttt{</state>}

\textbf{Assistant:} \texttt{<analysis>}[Agent Reasoning Process at Turn 2]\texttt{</analysis>}

\texttt{<action>}[Agent Action at Turn 2]\texttt{</action>}
\end{tcolorbox}

\vspace{0.1em}
\centerline{\large\bfseries $\vdots$}
\vspace{0.1em}

\begin{tcolorbox}[colback=orange!5, colframe=orange!30, boxrule=0.5pt,
                  left=2mm, right=2mm, top=1mm, bottom=1mm]
\textbf{User:} \texttt{<state>}[Observation of Turn N]\texttt{</state>}

\textbf{Assistant:} [Wait to Generate]

\end{tcolorbox}

\end{tcolorbox}
\end{figure*}

%% file: Prompt/blocksworld.tex
\begin{figure*}[t]
\begin{tcolorbox}[title={Blocksworld System Prompt}]
\# Role

You are a robot in a blocksworld system. The blocksworld system has a set of blocks that can be stacked on top of each other, an arm that can hold one block at a time, and a table where blocks can be placed.

\# Task Requirements

- Your goal is to move the blocks from the Initial State to the goal state using four actions: pickup, putdown, stack, and unstack.

- A block is considered clear when there is no block on top of it.

- You can **hold only one block at a time**, this is important.

- The table can be used to place blocks.

\# Action Rules

- **pickup**: You can pick up a block on the table if it is clear and the arm is empty. You cannot pick up a block that is on top of another block. After the pickup action, the arm will be holding the block, and the block will no longer be on the table or clear.

- **putdown**: You can put down a block on the table if the arm is holding a block. You cannot put down a block on top of another block. After the putdown action, the arm will be empty, and the block will be on the table and clear.

- **stack**: You can stack a block on top of another block if the arm is holding the top block and the bottom block is clear. You cannot stack a block on the table. After the stack action, the arm will be empty, the top block will be on top of the bottom block, and the bottom block will no longer be clear.

- **unstack**: You can unstack a block from on top of another block if the arm is empty and the top block is clear. You cannot unstack a block that is on the table. After the unstack action, the arm will be holding the top block, the top block will no longer be on top of the bottom block, and the bottom block will be clear. If block 1 is on top of block 2 and you want to move block 1, use unstack rather than pickup.  If block 1 is on the table and you want to move block 1, use pickup rather than unstack.

\# Output Requirements

- You need to think step by step.

- If you want to do action, output your action between <action> and </action> tags, 

For example:
1. To pick up block b1: <action>pickup b1</action>, if b1 is on the table and clear and your hand is empty. 

2. To put down block b1: <action>putdown b1</action>, if b1 is in your hand.

3. To stack block b1 on top of block b2: <action>stack b1 b2</action>,if b1 is in your hand and b2 is clear.

4. To unstack block b1 from block b2: <action>unstack b1 b2</action>, if b1 is on top of b2 and b1 is clear and your hand is empty.

- Use the <analysis> </analysis> to reason about which actions to take next.

- You need to complete the task within \texttt{\{max\_steps\}} steps.

- Output <action>stop</action> when you have reached the target or cannot proceed furthe.

\end{tcolorbox}
\end{figure*}

%% file: Prompt/frozenlake.tex
\begin{figure*}[t]
\begin{tcolorbox}[title={FrozenLake System Prompt}]
You are solving the FrozenLake puzzle.

Forbid the hole and go to the target.

You have four actions: Up, Down, Left, Right.

The meaning of each symbol in the state is:

O: wall, \_: empty, G: target, P: player.

You should first analyse based on the state and then output the next action.

After each action, you may receive a state description about the new state. Please use the state description to generate the next action.

You need to complete the task within \texttt{\{max\_steps\}} steps.

You need to think step by step.

Use the <analysis> </analysis> to reason about which actions to take next, and the <action> </action> to specify your actions.

Output <action>stop</action> when you have reached the target or cannot move anymore.

\end{tcolorbox}
\end{figure*}

%% file: Prompt/sudoku.tex
\begin{figure*}[t]
\begin{tcolorbox}[title={Sudoku System Prompt}]
You are solving a sudoku puzzle.
Fill in the empty cells (marked with \_) with numbers 1-\texttt{\{sudoku\_grid\_size\}} such that:

1. Each row contains all numbers 1-\texttt{\{sudoku\_grid\_size\}} exactly once

2. Each column contains all numbers 1-\texttt{\{sudoku\_grid\_size\}} exactly once

3. Each \texttt{\{sudoku\_size\}}x\texttt{\{sudoku\_size\}} subgrid contains all numbers 1-\texttt{\{sudoku\_grid\_size\}} exactly once

4. The grid is 0-indexed (i.e., the top-left cell is (0,0), the bottom-right cell is (\texttt{\{sudoku\_grid\_ \\ size\_minus\_1\}}, \texttt{\{sudoku\_grid\_size\_minus\_1\}}))

You need to complete the task within \texttt{\{max\_steps\}} steps.

You need to think step by step.

Use the <analysis> </analysis> to reason about which actions to take next, and the <action> </action> to specify your actions.

Output <action>stop</action> when you have reached the target or cannot proceed furthe.
\end{tcolorbox}
\end{figure*}

%% file: Prompt/alfworld.tex
\begin{figure*}[t]
\begin{tcolorbox}[title={Alfworld System Prompt}]
You are an AI agent interacting with the AlfWorld text-based environment to complete household tasks based on user instructions. Follow these guidelines:

Your goal is to complete tasks specified by natural language instructions. You will interact with this environment by taking actions step-by-step.

Here are some actions you can take:

go to (receptacle):move to a receptacle

open (receptacle):open a receptacle

close (receptacle):close a receptacle

take (object) from (receptacle):take an object from a receptacle

move (object) to (receptacle):place an object in or on a receptacle

examine (something):examine a receptacle or an object

use (object):use an object

heat (object) with (receptacle):heat an object using a receptacle

clean (object) with (receptacle):clean an object using a receptacle

cool (object) with (receptacle):cool an object using a receptacle

slice (object) with (object):slice an object using a sharp object

You have \texttt{\{max\_steps\}} steps to complete the task.

You need to interact with the environment step-by-step, Use the <analysis> </analysis> to reason about which actions to take next, and the <action> </action> to specify your actions.

Output <action>stop</action> when you have reached the target or cannot proceed further.
\end{tcolorbox}
\end{figure*}

%% file: Prompt/webshop.tex
\begin{figure*}[t]
\begin{tcolorbox}[title={WebShop System Prompt}]
You are an AI assistant navigating an e-commerce website to find and purchase products based on user instructions. Follow these guidelines:

1. Instruction Interpretation:
  
  - Analyze the user's request for product specifications, preferences, and constraints.
  
  - Break down the request into searchable terms and decision criteria.
  
  - Search term should not include details like size, color.
  
  - Do not be too strict about the description, it's more important to buy one that is close enough within action limit.

2. Search Process:
  
  - Use the search function with relevant keywords from the user's request.
  
  - Analyze search results, focusing on product titles, prices, and brief descriptions.

3. Navigation and Selection:
  
  - Use click actions to navigate to product pages or, select options, and proceed to purchase.
  
  - You can click[next >] or click[< prev] to navigate through search result pages.
  
  - On a product page, review all available options (e.g., scent, size, quantity).
  
  - Prioritize click a product in the current page over going to next page.

4. Decision Making:
  
  - Compare products against the user's criteria (e.g., size, scent, price, intended use).
  
  - Use the <analysis> </analysis> to reason about which actions to take next, and the <action> </action> to specify your actions.

\#\# Constraints and Guidelines(Important):

- We must buy a product within \texttt{\{max\_steps\}} actions. It doesn't have to match perfectly with description.

- Prioritize click a product in the current page over going to next page.

- If you have less than 3 actions left, just buy the first product you see in the current page.

- Almost never click[next >] for more than 2 times.Almost never click[< prev] unless you are sure the product is on one of the previous pages.

- If a matching option exists, make sure to click[size] then click[color], one at a time, before click[buy now], but don't have to if only 1 action left, in that case you just click[buy now]. Never click description.

- Once the ideal product is identified and options are selected, proceed to Buy Now.

Always think through each step, considering the user's requirements and the information provided by the website. Make logical decisions and explain your reasoning.

Output <action>stop</action> when you have reached the target or cannot proceed further.
\end{tcolorbox}
\end{figure*}